\documentclass[acmtog,nonacm]{acmart}

\usepackage{xcolor}
\usepackage{graphicx}
\usepackage{booktabs}
\usepackage{multirow}
\usepackage{calc}
\usepackage{enumitem}
\usepackage{algorithm}
\usepackage{algpseudocode}
\usepackage{amsmath}
\usepackage{xspace}
\providecommand{\checkmark}{\ensuremath{\surd}}

\newcommand{\sysname}[0]{4DAnyone}
\newcommand{\papertitle}{4DAnyone: Create Anyone in 4D from a Casual Monocular Video}
\newcommand{\resulttablerowstretch}{1.03}
\makeatletter
\newcommand{\maketitlesupplementary}[1]{%
    \hsize=\textwidth
    \@ACM@title@width=\hsize
    \setbox\mktitle@bx=\vbox{\noindent\@titlefont
        \parbox[t]{\@ACM@title@width}{\raggedright
            \@titlefont\noindent
            #1%
            \par\vskip\dimexpr\bigskipamount+1pt\relax
            \noindent{\huge Supplementary Material}%
        }%
        \par\vskip18pt}%
    \twocolumn[\box\mktitle@bx]%
}
\makeatother
\newcommand{\PAR}[1]{\vskip4pt \noindent{\bfseries #1~}}

\definecolor{tabfirst}{rgb}{1, 0.7, 0.7}
\definecolor{tabsecond}{rgb}{1, 0.85, 0.7}
\definecolor{tabthird}{rgb}{1, 1, 0.7}
\definecolor{tabgray}{rgb}{0.9, 0.9, 0.9}

\newcommand{\eg}{\emph{e.g.}\xspace}

\setcopyright{none}
\renewcommand\footnotetextcopyrightpermission[1]{}

\newcommand{\suppref}[1]{\ref{#1}}

\begin{document}

\title{\papertitle}

\author{Yudong Jin}
\ifdefined\supplementaryversion\else
\authornote{Equal contribution.}
\fi
\orcid{0009-0006-1887-5982}
\email{krahetx@gmail.com}
\affiliation{%
  \institution{State Key Lab of CAD\&CG, Zhejiang University}
  \city{Hangzhou}
  \country{China}}
\affiliation{%
  \institution{Robbyant}
  \city{Hangzhou}
  \country{China}}

\author{Tao Xie}
\ifdefined\supplementaryversion\else
\authornotemark[1]
\fi
\orcid{0009-0007-8644-5324}
\email{xbillowy@gmail.com}
\affiliation{%
  \institution{State Key Lab of CAD\&CG, Zhejiang University}
  \city{Hangzhou}
  \country{China}}

\author{Qihang Zhang}
\orcid{0000-0003-1784-8166}
\email{zqh10241024@gmail.com}
\affiliation{%
  \institution{Robbyant}
  \city{Beijing}
  \country{China}}
\affiliation{%
  \institution{Chinese University of Hong Kong}
  \city{Hong Kong}
  \country{China}}

\author{Zehong Shen}
\orcid{0000-0002-7232-793X}
\email{zhshen0917@gmail.com}
\affiliation{%
  \institution{Ant Group}
  \city{Hangzhou}
  \country{China}}

\author{Zhen Xu}
\orcid{0009-0002-6098-4198}
\email{zhenx@zju.edu.cn}
\affiliation{%
  \institution{State Key Lab of CAD\&CG, Zhejiang University}
  \city{Hangzhou}
  \country{China}}

\author{Yujun Shen}
\orcid{0000-0003-3801-6705}
\email{shenyujun0302@gmail.com}
\affiliation{%
  \institution{Robbyant}
  \city{Hangzhou}
  \country{China}}

\author{Hujun Bao}
\orcid{0000-0002-2662-0334}
\email{bao@cad.zju.edu.cn}
\affiliation{%
  \institution{State Key Lab of CAD\&CG, Zhejiang University}
  \city{Hangzhou}
  \country{China}}

\author{Xiaowei Zhou}
\ifdefined\supplementaryversion\else
\authornote{Corresponding authors.}
\fi
\orcid{0000-0003-1926-5597}
\email{xwzhou@zju.edu.cn}
\affiliation{%
  \institution{State Key Lab of CAD\&CG, Zhejiang University}
  \city{Hangzhou}
  \country{China}}

\author{Yinghao Xu}
\ifdefined\supplementaryversion\else
\authornotemark[2]
\fi
\orcid{0000-0003-2696-9664}
\email{justimyhxu@gmail.com}
\affiliation{%
  \institution{Hong Kong University of Science and Technology}
  \city{Hong Kong}
  \country{China}}
\affiliation{%
  \institution{Robbyant}
  \city{Beijing}
  \country{China}}

\newcommand{\pdfauthors}{%
  Yudong Jin, Tao Xie, Qihang Zhang, Zehong Shen, Zhen Xu, Yujun Shen,
  Hujun Bao, Xiaowei Zhou, and Yinghao Xu}

\makeatletter
\authorsaddresses{\scriptsize\@mkauthorsaddresses}
\makeatother

\begin{teaserfigure}
\centering
\includegraphics[width=\textwidth]{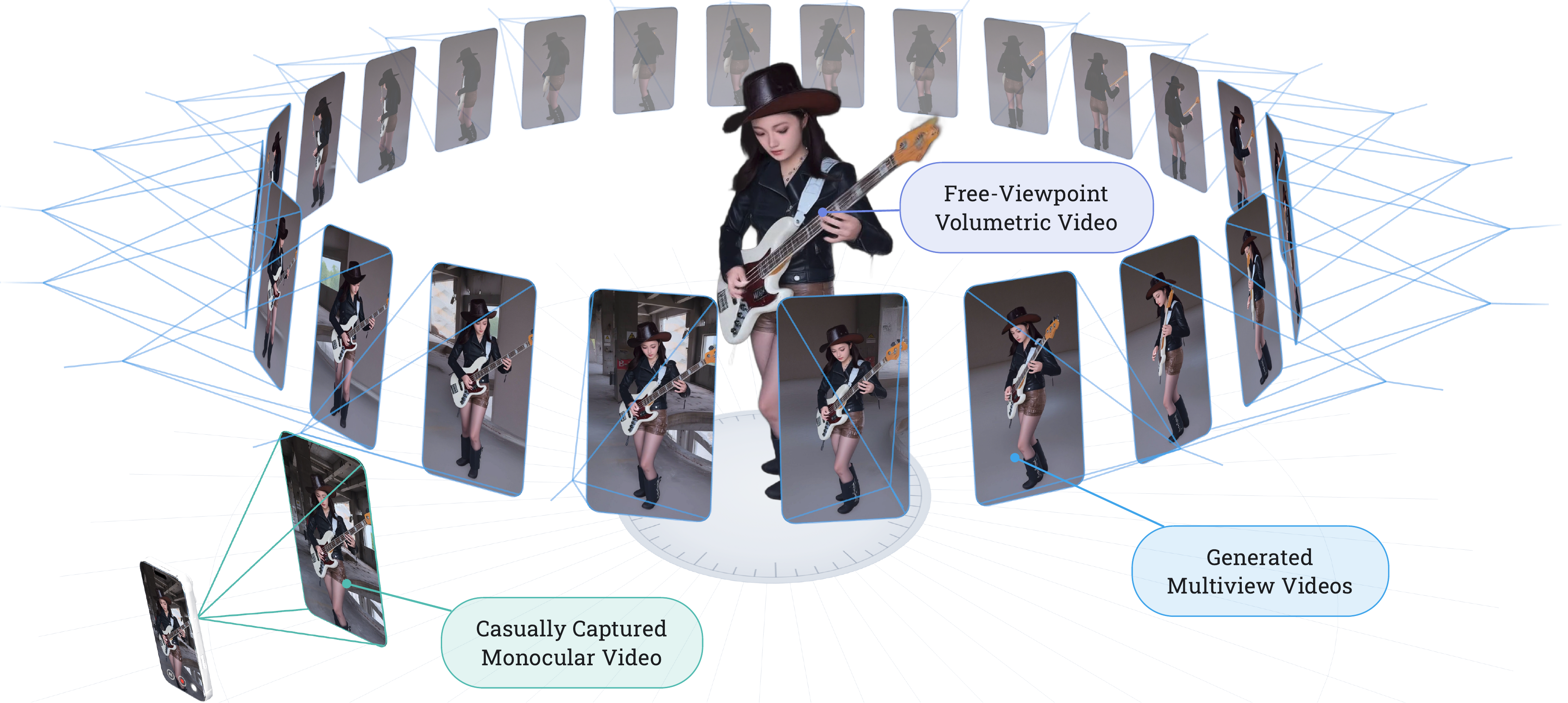}
\caption{Given a casually captured monocular video, typically with mild camera motion and unknown camera intrinsics and poses, \sysname{} generates multiview-consistent human videos, enabling 4DGS reconstruction rendered from free viewpoints. See the project page and Fig.~\ref{fig:wild_demo} for more results.}
\Description{One continuous 3D scene: a smartphone at the lower left films a bass-playing subject, a ring of twenty-four generated camera views surrounds the stage, and the reconstructed subject stands at the center on a turntable, rendered from a free viewpoint.}
\label{fig:teaser}
\end{teaserfigure}

\begin{abstract}
We present \sysname{}, a framework for reconstructing 4D humans from an uncalibrated monocular video by generating reconstruction-grade multiview-consistent videos and lifting them into 4D Gaussian Splatting (4DGS).
Existing camera-controlled video diffusion models synthesize plausible novel-view videos but fail to maintain consistency when scaled to the tens of target views required for 4DGS reconstruction.
We identify this failure as a bounded-attention-context problem: when target views exceed the capacity of a single DiT forward pass, they must be split into groups, exposing two coupled bottlenecks.
On the reference-context side, conditioning on all previously generated views grows as $O(N)$, weakening cross-view appearance guidance.
On the target-context side, disjoint groups cannot directly exchange information, causing global structural drift.
\sysname{} addresses both bottlenecks with two complementary designs: Reference Context Packing (RCP) compresses growing reference views into a fixed-length mixed-resolution context with $O(1)$ reference-context complexity, while Target Context Routing (TCR) rotates target-view groupings during denoising to share context across groups at high-noise steps and stabilize details at low-noise steps.
We further build the MVGameHuman dataset using our in-house game engine and combine it with light-stage and in-the-wild video datasets for training.
Experiments on DNA-Rendering and DyMVHumans show that \sysname{} outperforms prior methods in both novel-view video quality and downstream 4DGS reconstruction, with robust in-the-wild generalization.
See our project page for video results and source code: \href{https://4danyone.github.io}{\textcolor[RGB]{53,103,198}{\nolinkurl{https://4danyone.github.io}}}.
\end{abstract}


\maketitle

\clearpage
\section{Introduction}
\label{sec:intro}

Reconstructing 4D humans renderable from arbitrary viewpoints is important for embodied AI, immersive content creation, and virtual reality.
Recent 4D Gaussian Splatting (4DGS) methods~\cite{kerbl2023gaussian, wu20244dgs, xu2024longvolcap, wang2025freetimegs} enable real-time, photorealistic dynamic-scene rendering, yet constructing such representations still requires dense multi-view video from calibrated, static camera arrays~\cite{cheng2023dna}.
This raises a natural question: \emph{can we reconstruct a 4D human from an uncalibrated monocular video?}

A straightforward attempt is to reconstruct 4D humans directly from monocular video~\cite{yu2023monohuman, hu2024gaussianavatar, hu2024gauhuman}.
However, these methods typically require known camera parameters and struggle to infer occluded body regions and recover high-fidelity appearance.
A more promising pipeline is to first generate novel-view videos with a camera-controlled video diffusion model and then perform 4DGS reconstruction.
Existing camera-controlled video generation methods, including implicit camera-conditioned models~\cite{bai2025recammaster,wu2025cat4d} and explicit geometry-conditioned models~\cite{ren2025gen3c,yu2025trajectorycrafter}, can synthesize visually plausible novel-view videos under prescribed camera motion.
However, when extended to the tens of target-view videos required for 4DGS reconstruction, these methods fail to maintain cross-view consistency, leading to appearance inconsistency and structural drift that degrade downstream 4D reconstruction.
The bottleneck is therefore not only view control, but \emph{how to maintain cross-view consistency at reconstruction scale}.

Why does consistency break at reconstruction scale? We argue the root cause is an architectural constraint: \emph{the attention context of a single DiT forward pass is bounded by practical memory and compute budgets}.
When the number of target views $N$ exceeds this capacity, the target context must be split into groups for tractable denoising, creating two consistency bottlenecks.
On the \emph{reference context side}, each group ideally conditions on all previously generated views, but the reference context length grows as $O(N)$ and quickly exceeds capacity.
On the \emph{target context side}, target contexts of different groups are disjoint, so cross-group views cannot directly attend to one another, leading to cross-group structural drift.
Existing camera-controlled video generation methods do not jointly resolve these two bottlenecks: limiting the reference context weakens appearance guidance, while independently denoising target groups prevents cross-group structural communication.

We present \sysname{}, which addresses both bottlenecks to enable reconstruction-grade multiview-consistent video generation.
First, \emph{Reference Context Packing (RCP)} exploits the redundancy of cross-view appearance: a mixed-resolution reference context often suffices to preserve global layout while retaining fine appearance details.
It keeps reference conditioning scalable by compressing the growing set of generated reference views into a fixed-length context, reducing the reference-context complexity from $O(N)$ to $O(1)$.
Second, \emph{Target Context Routing (TCR)} exploits the temporal structure of diffusion: high-noise steps tend to govern global structure, while low-noise steps mainly refine local details.
During high-noise denoising, TCR rotates target-view groupings to enable cross-group context sharing and propagate global structure.
During low-noise denoising, it fixes adjacent-view groups to stabilize fine details.
For geometric guidance, we follow an accuracy-over-density principle: instead of relying on dense metric depth, which is difficult to estimate reliably from unconstrained videos, we use 3D skeletons that modern human mesh recovery methods can reliably estimate from monocular input~\cite{shen2024gvhmr,yang2026sam3dbody}.
Accordingly, we adopt \emph{3D-aware skeleton conditioning} that encodes depth-buffered skeleton renderings to resolve the inherent 2D pose ambiguity.
We further build the \emph{MVGameHuman} dataset using our in-house game engine and combine it with light-stage and in-the-wild video datasets for training, enabling robust in-the-wild generalization.

In summary, our contributions are:
\begin{itemize}[nosep,leftmargin=*]
    \item We present \sysname{}, a 4D human video generation framework that achieves reconstruction-grade multiview consistency from monocular videos.
    \item We propose Reference Context Packing (RCP), which compresses the growing reference context from $O(N)$ to $O(1)$ while preserving cross-view appearance guidance.
    \item We propose Target Context Routing (TCR), which dynamically routes target-view groupings during denoising to enable cross-group context sharing and reduce cross-group structural drift.
\end{itemize}

\section{Related Work}
\label{sec:related_work}

\PAR{Camera-controlled video generation.}
Camera control in video diffusion broadly follows implicit or explicit conditioning.
Implicit methods encode camera information through learned representations without hard geometric constraints.
CameraCtrl~\cite{he2025cameractrl,he2025cameractrlii} injects pixel-wise Pl\"ucker rays into video diffusion U-Nets; MotionCtrl~\cite{wang2024motionctrl} separates camera and object motion using rotation-translation matrices; VD3D~\cite{bahmani2024vd3d} uses spatiotemporal camera embeddings in diffusion transformers; and CamCo~\cite{xu2024camco} combines Pl\"ucker conditioning with epipolar attention.
ReCamMaster~\cite{bai2025recammaster} concatenates source-video tokens with target trajectories to re-render novel views, while CAT4D~\cite{wu2025cat4d} extends multi-view diffusion to dynamic content.
Although these methods generalize well across diverse scenes, their latent camera control lacks hard geometric constraints and can drift under large viewpoint changes, limiting downstream 3D reconstruction.
Explicit methods improve accuracy with dense 3D geometry: Gen3C~\cite{ren2025gen3c} projects depth-derived point clouds for world-consistent generation, TrajectoryCrafter~\cite{yu2025trajectorycrafter} warps reference frames using estimated depth for trajectory redirection, and WVD~\cite{zhang2025world} jointly models RGB and XYZ frames to unify camera control with geometric reconstruction.
However, these methods depend on dense metric depth or 3D coordinates that are difficult to estimate reliably from in-the-wild videos~\cite{xu2025depthfoundation}, and remain limited for dynamic scenes or large viewpoint changes.

For humans, body pose offers an alternative geometric signal.
ControlNet~\cite{zhang2023controlnet}, MagicAnimate~\cite{xu2024magicanimate}, Animate Anyone~\cite{hu2024animateanyone}, UniAnimate~\cite{wang2025unianimate}, and Wan-Animate~\cite{cheng2025wananimate} use 2D pose skeletons for pose-driven animation, while 3DiMo~\cite{fang20263dimo} employs implicit motion encoding for view-adaptive generation.
These methods target motion transfer rather than reconstruction-grade novel view synthesis.
\sysname{} uses sparse, precise 3D skeletons for explicit conditioning, achieving reconstruction-grade multiview consistency without estimating dense depth or source-camera parameters.

\PAR{Multi-view diffusion models.}
\label{sec:mvdiffusion}
Multi-view diffusion generates geometrically consistent views for generation-then-reconstruction.
Zero-1-to-3~\cite{liu2023zero123} pioneered viewpoint-conditioned image generation from a single image.
MVDream~\cite{shi2024mvdream}, CAT3D~\cite{gao2024cat3d}, SV3D~\cite{voleti2024sv3d}, and Zero123++~\cite{shi2023zero123++} further advance multi-view generation for static 3D reconstruction.
SV4D~\cite{xie2024sv4d} and CAT4D~\cite{wu2025cat4d} extend this paradigm to dynamic content through multi-view video diffusion, while Diffuman4D~\cite{jin2025diffuman4d} further targets 4D human reconstruction but requires synchronized sparse-view videos with known camera parameters.
Scaling to many reconstruction views remains difficult: GPU memory limits the views per forward pass, whereas independent batches cause cross-batch inconsistency.
CAT3D selects sparse anchor views as context for each batch, but the subset selection discards appearance information.
CAT4D and Diffuman4D use sliding-window denoising with overlap aggregation, yet cross-window drift persists at large view counts.
From monocular input alone, \sysname{} uses two complementary designs.
Reference Context Packing packs all reference views into a fixed-length context at $O(1)$ cost, preserving rich appearance cues without lossy anchor selection.
Target Context Routing regroups views at high noise to propagate global structure and reduce cross-group drift, then fixes adjacent groups at low noise for detail refinement, enabling consistent generation across many viewpoints.

\PAR{4D human avatar reconstruction.}
Traditional approaches reconstruct dynamic human avatars from dense multi-view captures using neural radiance fields~\cite{peng2021neuralbody, mildenhall2021nerf, lin2023im4d} or Gaussian splatting~\cite{kerbl2023gaussian, wu20244dgs, yang2024realtime4dgs, xu20244k4d, duan20244drotor, xu2024longvolcap, jiang2024dualgs, wang2025freetimegs, jiang2025taogs}, but require expensive multi-camera setups such as the 48-camera rig of DNA-Rendering~\cite{cheng2023dna}.
Monocular methods~\cite{weng2022humannerf, jiang2023instantavatar, yu2023monohuman, hu2024gaussianavatar, qian20243dgsavatar, hu2024gauhuman, hu2024expressive} remove this dependency by fitting neural representations to single-view video with parametric body priors~\cite{smpl, pavlakos2019expressive}, yet they cannot reliably hallucinate appearance in unobserved regions, placing an inherent ceiling on visual quality.
UP2You~\cite{cai2025up2you} reconstructs 3D clothed portraits from unconstrained in-the-wild photos by rectifying unstructured inputs into clean multi-view images via a pose-correlated feature aggregation module, but is limited to static reconstruction.
MV-Performer~\cite{zhi2025mvperformer} addresses human novel view synthesis with explicit geometric conditioning but remains within a reconstruction-only paradigm without generative hallucination.
\sysname{} takes a generation-assisted approach: it synthesizes multi-view observations from a monocular video via skeleton-conditioned diffusion, then applies standard 4DGS pipelines~\cite{wu20244dgs, wang2025freetimegs} for high-fidelity 4D human reconstruction without specialized hardware.

\section{Method}
\label{sec:method}

\subsection{Overview}
\label{sec:overview}

\begin{figure*}[t]
\centering
\includegraphics[width=\textwidth]{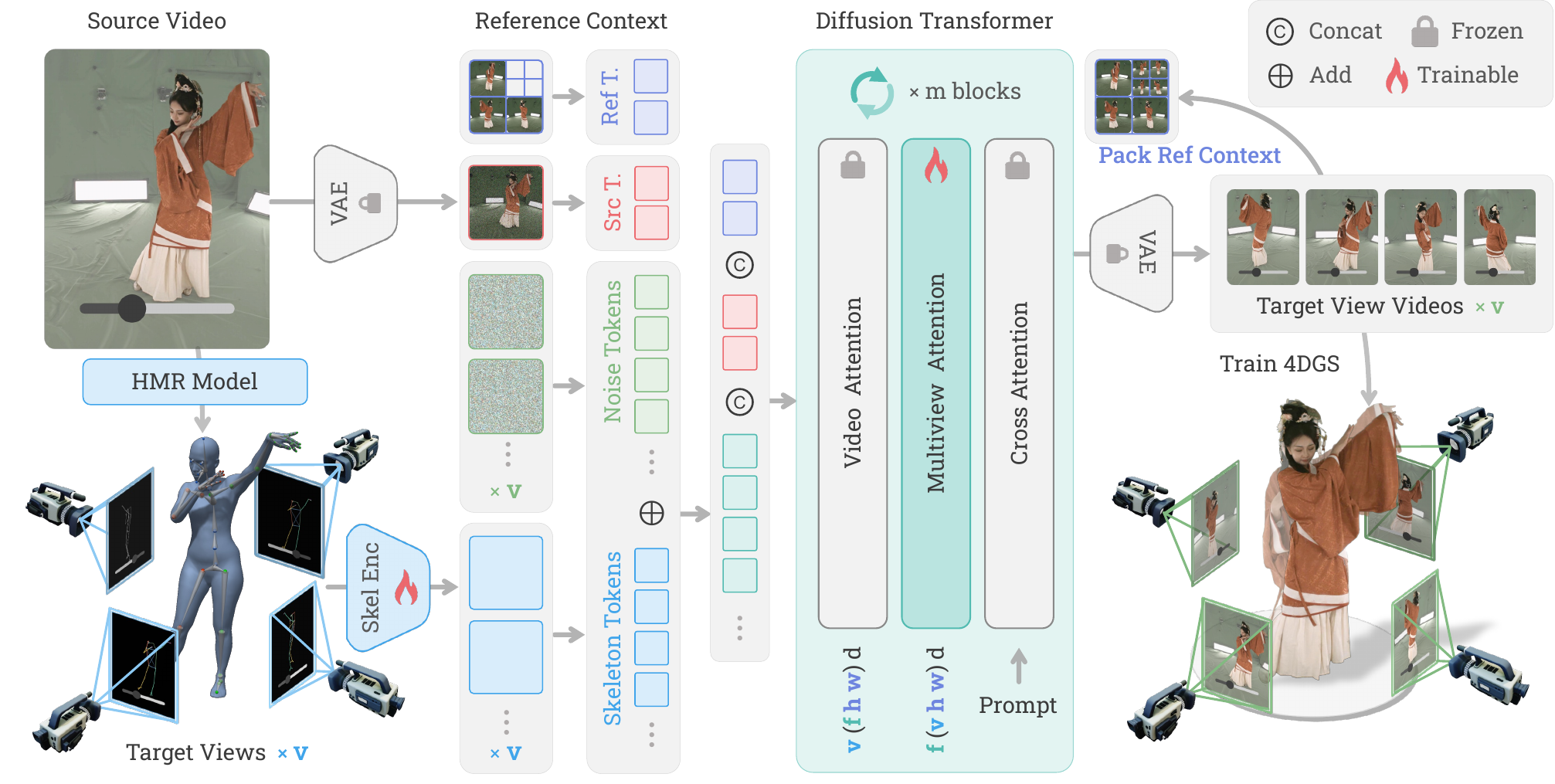}

\caption{\textbf{Overview of \sysname{}.} Given a source video, an HMR model (GVHMR~\cite{shen2024gvhmr}) estimates a 3D skeleton sequence, which is rendered into depth-buffered skeleton videos for $v$ target views.
A 3D-aware skeleton encoder produces residual skeleton tokens that are added to the noisy target latents.
The source tokens, RCP reference tokens, and skeleton-conditioned target tokens are concatenated along the view dimension and processed by the DiT with video attention, multiview attention, and text cross-attention.
Generated target videos are packed back into the RCP context for subsequent generation rounds, while TCR improves consistency during grouped target-view generation.
The final multi-view videos are used to train a 4DGS model with FreeTimeGS~\cite{wang2025freetimegs}.}
\Description{Pipeline diagram. A source video is converted into skeleton-conditioned target-view tokens. Reference Context Packing supplies compact reference tokens, Target Context Routing exchanges information among target-view groups, and the generated multiview videos are reconstructed as a 4D Gaussian Splatting model.}
\label{fig:pipeline}
\end{figure*}

Given a monocular source video $\mathbf{V}_{\text{src}} \in \mathbb{R}^{F \times 3 \times H \times W}$ with unknown camera intrinsics and poses, \sysname{} generates $v$ human videos $\{\mathbf{V}_i\}_{i=1}^{v}$ with reconstruction-grade consistency at prescribed static viewpoints distributed around the subject for high-fidelity 4DGS reconstruction.
As shown in Fig.~\ref{fig:pipeline}, HMR estimates a 3D skeleton sequence, which we render into depth-buffered skeleton videos at the target viewpoints.
A 3D-aware skeleton encoder injects these cues into target latents (Sec.~\ref{sec:skeleton}), which the DiT denoises conditioned on the source video, skeletons, and Reference Context Packing (RCP).
For tens of views, Target Context Routing (TCR) exchanges context among target groups and reduces structural drift under a fixed per-group budget (Sec.~\ref{sec:scale}).
FreeTimeGS~\cite{wang2025freetimegs} then reconstructs the 4DGS model from the generated videos.

\subsection{3D-Aware Skeleton Conditioning}
\label{sec:skeleton}

Prior geometry-conditioned methods~\cite{ren2025gen3c,yu2025trajectorycrafter} rely on dense geometric signals (\eg depth maps and camera parameters) for viewpoint control.
However, such signals are difficult to estimate reliably from in-the-wild videos, and errors in these estimates can introduce conflicting geometric constraints that cause multiview generation to diverge.
Our key observation is \emph{accuracy over density}: for reconstruction-grade multiview consistency, the accuracy of geometric signals is more critical than their density.
Accordingly, we use 3D skeletons as sparse but reliable geometric guidance: they provide accurate structural cues, leave appearance details to the video model, and can be robustly recovered from monocular videos using modern human mesh recovery methods~\cite{shen2024gvhmr}.
Despite their sparsity, the video model can still learn precise spatial correspondences between skeleton keypoints and generated human content, maintaining consistency in body structure, clothing, facial expressions, and other details.

\PAR{Depth-buffered skeleton rendering.}
A naive 2D skeleton rendering suffers from inherent pose ambiguity: occlusions between body parts are lost, so opposite front-back configurations (\eg an arm in front of or behind the torso) produce the same rendering, causing inconsistent results across viewpoints.
We therefore rasterize the 3D skeleton with a pixelwise z-buffer so that nearer body parts correctly occlude farther ones, upgrading the ambiguous 2D skeleton to an occlusion-aware rendering without extra input channels.

\PAR{Keypoint selection.}
To improve the reliability of skeleton conditioning, we use a compact 40-keypoint subset of the 308-keypoint Goliath vocabulary~\cite{khirodkar2026sapiens2}, retaining 17 body, 6 foot, and 10 palm-level hand keypoints (5 knuckles per hand) plus 7 auxiliary neck, shoulder, and elbow landmarks, while excluding the 238 facial keypoints and 30 finger joints.
Facial expressions and fine hand details are instead learned from the source-video reference, avoiding artifacts from noisy fine-grained keypoint detections.

\PAR{Skeleton encoder.}
The skeleton encoder $g_\phi$ injects the skeleton condition into the noisy latent tokens as a DiT-resolution residual:
\[
\tilde{\mathbf{z}}^t_i
= \mathbf{z}^t_i + g_\phi\!\left(\mathbf{S}_i\right),
\]
where $\mathbf{S}_i$ denotes the depth-buffered skeleton video for target view $i$, and $\mathbf{z}^t_i$ denotes the noisy latent tokens at timestep $t$.
The final projection layer of $g_\phi$ is zero-initialized for stable training from the pretrained DiT weights.
See Supp.~\suppref{sec:supp_model} for details.

\subsection{Scalable Multiview Consistency}
\label{sec:scale}

\begin{figure*}[t]
\centering
\includegraphics[width=\textwidth]{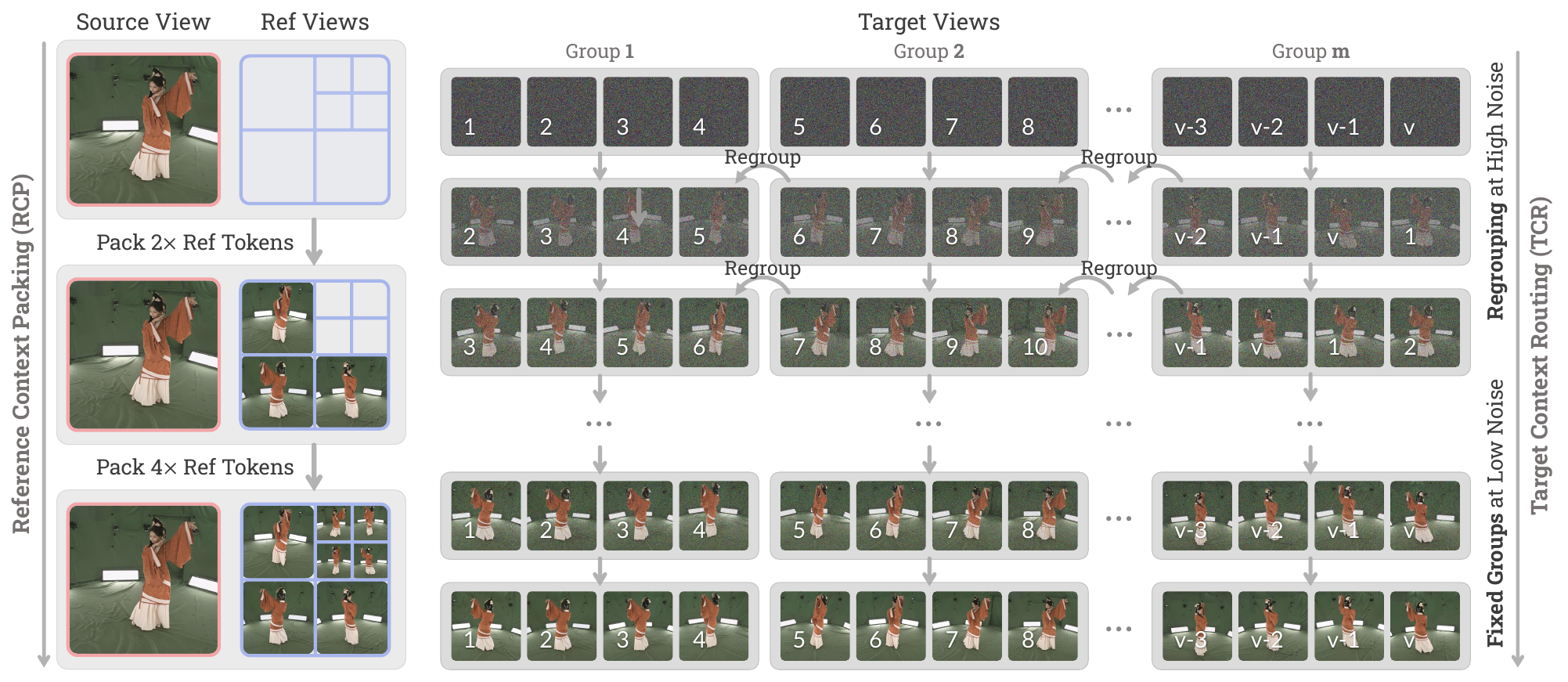}
\caption{\textbf{Progressive inference with Reference Context Packing and Target Context Routing.} We first generate reference videos and pack them with the source video into a fixed RCP context. During target-view generation, this fixed RCP context is shared by all groups. TCR partitions $v$ target views into $m$ four-view groups, cyclically regroups them at high noise levels to propagate global structure, and fixes adjacent groups at low noise levels for stable detail refinement.}
\Description{Three-stage inference diagram. Initial views become a fixed multiscale reference context; target views are cyclically regrouped at high noise to share global structure and arranged into fixed adjacent groups at low noise to refine local details.}

\label{fig:rcp_tcr}
\end{figure*}

High-fidelity 4DGS reconstruction requires tens of target-view videos (\eg 16), but joint denoising is prohibitively expensive.
Splitting views into groups introduces two coupled consistency bottlenecks.
On the conditioning side, each group needs appearance references that cover back and side views unseen in the monocular input.
On the generation side, independently denoised groups cannot exchange global structural information and may drift across groups.
As shown in Fig.~\ref{fig:rcp_tcr}, \emph{Reference Context Packing} supplies scalable appearance context, and \emph{Target Context Routing} propagates structure across groups.

\PAR{Reference Context Packing.}
\label{sec:rcp}
Across multiple target-view generation rounds, earlier outputs can serve as appearance references for later rounds to enhance multiview consistency.
However, these references contain substantial cross-view redundancy, especially among nearby viewpoints.
Directly appending every generated view as full DiT context is therefore inefficient, as the context length and computation grow linearly with the number of references.
Inspired by FramePack~\cite{zhang2025framepack}, RCP uses multi-scale patchify layers to pack the growing set of target-view references into a fixed set of reference token slots.
Specifically, while the standard Wan2.2 patchify layer uses kernel and stride $(1, 2, 2)$, RCP uses a patchify layer $\mathcal{P}_r$ with kernel and stride $(1, 2r, 2r)$ for compression ratio $r$, producing $\frac{1}{r^2}$ as many tokens as the standard layer.
The packed reference tokens are assembled along the spatial dimensions and concatenated with the source and target tokens along the view dimension.
Because the token-slot budget is fixed, RCP provides scalable appearance guidance with $O(1)$ overhead as references accumulate.

In practice, we use $r \in \{2, 4\}$ for RCP references and denote the resulting fixed RCP context as
\[
\mathcal{C}_{\text{R}} =
[\mathcal{P}_1(\mathbf{V}_{\text{src}}),
\{\mathcal{P}_2(\mathbf{V}_{a_j})\}_{j=1}^{3},
\{\mathcal{P}_4(\mathbf{V}_{b_j})\}_{j=1}^{4}].
\]
During training, we randomly sample $\mathbf{V}_{\text{src}}$, $\{\mathbf{V}_{a_j}\}$, and $\{\mathbf{V}_{b_j}\}$ from each training sequence.
To support progressive inference, we apply dropout to $\mathcal{C}_{\text{R}}$ by zeroing out either $\{\mathcal{P}_4(\mathbf{V}_{b_j})\}_{j=1}^{4}$ or both $\{\mathcal{P}_2(\mathbf{V}_{a_j})\}_{j=1}^{3}$ and $\{\mathcal{P}_4(\mathbf{V}_{b_j})\}_{j=1}^{4}$, with probability 0.1 each.

At inference time, we first generate reference views for RCP in two rounds.
For each round, we select target viewpoints by farthest-point sampling: starting from the source view and existing reference views, it iteratively adds the candidate with the largest minimum angular distance to the selected views, improving viewpoint coverage.
Round~1 generates four reference videos $\{\mathbf{V}_{j}\}_{j=1}^{4}$ conditioned only on $\mathcal{P}_1(\mathbf{V}_{\text{src}})$, and Round~2 generates four additional reference videos conditioned on $[\mathcal{P}_1(\mathbf{V}_{\text{src}}), \{\mathcal{P}_2(\mathbf{V}_{a_j})\}_{j=1}^{3}]$, where $\{\mathbf{V}_{a_j}\}_{j=1}^{3} \subset \{\mathbf{V}_{j}\}_{j=1}^{4}$.

After these reference views are generated, we generate all target views in four-view groups using a fixed RCP context $\mathcal{C}_{\text{R}}$ built from them.
Specifically, $\{\mathbf{V}_{a_j}\}_{j=1}^{3}$ is selected from Round~1 and $\{\mathbf{V}_{b_j}\}_{j=1}^{4}$ is selected from Round~2.
The denoising process for these groups is handled by TCR, described next.

\PAR{Target Context Routing.}
\label{sec:tcr}
RCP addresses the conditioning bottleneck by giving each target group scalable appearance references.
However, target groups are still denoised independently with no information exchange, so global structure can drift across groups despite shared appearance references.
TCR addresses this generation-side bottleneck by rotating target-view groupings during inference, so information can propagate across groups under the same per-group memory budget.

Our observation is that global structure is established at high noise levels, where isolated target groups are most likely to diverge in structure, eventually leading to inconsistent multiview appearance.
Accordingly, TCR divides inference into two phases based on a switching timestep $t_s$:
\begin{itemize}[nosep,leftmargin=*]
    \item \emph{High-noise phase} ($t > t_s$): At each step, we cyclically shift the ordered view indices by the step index and repartition them into four-view groups, letting views exchange context over time and propagate global structure under the same per-group budget.
    \item \emph{Low-noise phase} ($t \leq t_s$): We fix adjacent four-view groups so neighboring views jointly refine details, stabilizing appearance and cross-view transitions.
\end{itemize}

Algorithm~\ref{alg:tcr} summarizes TCR; at each denoising step, groups can be processed sequentially or in parallel across multiple GPUs.

\begin{algorithm}[t]
\caption{Target Context Routing}
\label{alg:tcr}
\small
\begin{algorithmic}[1]
\Require Target viewpoints $\mathcal{V}$, fixed RCP context $\mathcal{C}_{\text{R}}$ built from the generated reference views, switching timestep $t_s$, denoising schedule $\{t_T, \ldots, t_0\}$, skeleton conditions $\{\mathbf{S}_i\}_{i \in \mathcal{V}}$
\State Initialize $\mathbf{z}_i^T \sim \mathcal{N}(0, \mathbf{I})$ for all views $i \in \mathcal{V}$
\For{$n = T, T{-}1, \ldots, 1$}
    \If{$t_n > t_s$} \Comment{High-noise: rotate groups}
        \State $\mathcal{G} \gets \textsc{RotatingGroups}(\mathcal{V},\; 4,\; n)$
    \Else \Comment{Low-noise: fix adjacent groups}
        \State $\mathcal{G} \gets \textsc{AdjacentGroups}(\mathcal{V},\; 4)$
    \EndIf
    \For{each group $G \in \mathcal{G}$}
        \State Jointly denoise $\{\mathbf{z}_i^n\}_{i \in G}$ conditioned on $\mathcal{C}_{\text{R}}$ and $\{\mathbf{S}_i\}_{i \in G}$ to obtain $\{\mathbf{z}_i^{n-1}\}_{i \in G}$
    \EndFor
\EndFor
\State Decode $\{\mathbf{z}_i^0\}_{i \in \mathcal{V}}$ into target-view videos $\{\mathbf{V}_i\}_{i \in \mathcal{V}}$
\State \Return $\{\mathbf{V}_i\}_{i \in \mathcal{V}}$
\end{algorithmic}
\end{algorithm}

\subsection{Training Protocol}
\label{sec:training}

\PAR{Multi-stage training.}
We adopt a three-stage curriculum to progressively build the model's capabilities:

\emph{Stage~1} trains on foreground-only DNA-Rendering videos to learn skeleton-conditioned camera control.
To decouple pose from appearance, we sample the source-view and target-view clips from different temporal windows with 20\% probability.
This encourages the model to follow the target skeleton while preserving appearance from the source video.

\emph{Stage~2} extends training to all multi-view datasets without foreground masking.
Prior methods~\cite{jin2025diffuman4d} train on masked human videos to avoid overfitting to uniform green-screen backgrounds, but this strategy introduces two drawbacks: foreground mask boundaries cause edge noise and multi-view inconsistency, while the absence of backgrounds prevents the model from learning lighting and shadow cues.
This stage addresses both issues.

\emph{Stage~3} adds monocular datasets to improve in-the-wild generalization.
In this stage, we remove finger keypoints from the skeleton input, encouraging the model to infer hand details from the source video rather than relying on noisy finger-keypoint detections.

\PAR{Loss functions.}
We train with a latent flow-matching loss and a perceptual reconstruction loss:
\[
\mathcal{L} = \mathcal{L}_{\text{latent}} + \lambda \mathcal{L}_{\text{LPIPS}},
\]
where $\mathcal{L}_{\text{latent}}$ is the standard MSE flow-matching loss in latent space.
We compute $\mathcal{L}_{\text{LPIPS}}$~\cite{zhang2018lpips} on decoded frames with $\lambda=0.25$ to mitigate artifacts caused by the high spatial compression of the Wan2.2 VAE.
To reduce GPU memory consumption, we evaluate LPIPS on body-part-aware semantic crops. See Supp.~\suppref{sec:supp_training} for details.

\section{Experiments}
\label{sec:experiments}

\subsection{Experimental Setup}
\label{sec:exp_setup}

\PAR{Dataset.}
Table~\ref{tab:dataset} summarizes our human-centric training data.
MVGameHuman and SynCamVideo~\cite{bai2025syncammaster} provide multi-view videos with diverse backgrounds, DNA-Rendering~\cite{cheng2023dna} provides real captures, and monocular Pexels and TedTalk improve in-the-wild generalization.
MVGameHuman is captured using our in-house game engine (Supp.~\suppref{sec:supp_dataset}).

We detect 2D keypoints using Sapiens2-1B~\cite{khirodkar2026sapiens2}.
For multi-view datasets, 3D skeletons are obtained via cross-view triangulation and reprojected to each camera view for depth-buffered skeleton rendering.
For monocular datasets, we sample per-keypoint depth from the Sapiens2 pointmap prediction to drive the z-buffer, and filter out sequences with large camera motion.

\begin{table}[t]
\centering
\caption{\textbf{Training dataset statistics.}}
\label{tab:dataset}
\small
\setlength{\tabcolsep}{0pt}
\begin{tabular*}{\columnwidth}{@{\extracolsep{\fill}}lccccc@{}}
\toprule
Dataset & Videos & Cameras & Actors & Resolution & Type \\
\midrule
MVGameHuman & 38k & 24 & 318 & 2560${\times}$1440 & Multi-view \\
SynCamVideo & 34k & 10 & 66 & 1280${\times}$1280 & Multi-view \\
DNA-Rendering & 51k & 48 & 548 & 2048${\times}$2048 & Multi-view \\
TedTalk & 42k & 1 & 413 & 2160${\times}$1620 & Monocular \\
Pexels & 20k & 1 & 1,411 & 3840${\times}$2160 & Monocular \\
\bottomrule
\end{tabular*}
\end{table}

\begin{table*}[t]
\centering
\caption{\textbf{Quantitative comparison on DNA-Rendering and DyMVHumans.} We evaluate 4DGS reconstruction, generated video consistency, and generated video reconstruction. Best results are in bold. $\dagger$ denotes our fine-tuned version.}
\label{tab:comparison}
\small
\setlength{\tabcolsep}{0pt}
\renewcommand{\arraystretch}{\resulttablerowstretch}
\begin{tabular*}{0.92\textwidth}{@{\extracolsep{\fill}}llcccccc@{}}
\toprule
& \multirow{2}{*}[-0.75ex]{Method}
& \multicolumn{3}{c}{DNA-Rendering}
& \multicolumn{3}{c}{DyMVHumans} \\
\cmidrule(lr){3-5} \cmidrule(lr){6-8}
& & PSNR$\uparrow$ & SSIM$\uparrow$ & LPIPS$\downarrow$
  & PSNR$\uparrow$ & SSIM$\uparrow$ & LPIPS$\downarrow$ \\
\midrule
\multirow{4}{*}{\shortstack[l]{Gen. Video\\Consistency}}
& MV-Performer~\cite{zhi2025mvperformer}        & 21.25 & 0.799 & 0.222 & 19.98 & 0.797 & 0.182 \\
& TrajectoryCrafter~\cite{yu2025trajectorycrafter} & 13.56 & 0.641 & 0.331 & 15.19 & 0.769 & 0.204 \\
& ReCamMaster$^\dagger$~\cite{bai2025recammaster}          & 21.47 & 0.806 & 0.210 & 21.94 & 0.833 & 0.142 \\
& Ours                              & \textbf{24.33} & \textbf{0.862} & \textbf{0.163} & \textbf{24.48} & \textbf{0.862} & \textbf{0.109} \\
\midrule
\multirow{4}{*}{\shortstack[l]{4DGS\\Reconstruction}}
& MV-Performer~\cite{zhi2025mvperformer}        & 20.38 & 0.826 & 0.191 & 18.69 & 0.787 & 0.158 \\
& TrajectoryCrafter~\cite{yu2025trajectorycrafter} & 14.81 & 0.718 & 0.331 & 15.11 & 0.748 & 0.217 \\
& ReCamMaster$^\dagger$~\cite{bai2025recammaster}          & 20.55 & 0.807 & 0.214 & 19.86 & 0.795 & 0.159 \\
& Ours                              & \textbf{24.15} & \textbf{0.863} & \textbf{0.159} & \textbf{23.28} & \textbf{0.846} & \textbf{0.117} \\
\midrule
\multirow{4}{*}{\shortstack[l]{Gen. Video\\Reconstruction}}
& MV-Performer~\cite{zhi2025mvperformer}        & 19.33 & 0.803 & 0.204 & 14.36 & 0.731 & 0.230 \\
& TrajectoryCrafter~\cite{yu2025trajectorycrafter} & 13.68 & 0.698 & 0.358 & 14.11 & 0.725 & 0.247 \\
& ReCamMaster$^\dagger$~\cite{bai2025recammaster}          & 20.74 & 0.809 & 0.204 & 19.18 & 0.778 & 0.168 \\
& Ours                              & \textbf{23.69} & \textbf{0.850} & \textbf{0.165} & \textbf{21.03} & \textbf{0.808} & \textbf{0.143} \\
\bottomrule
\end{tabular*}
\end{table*}

\begin{table}[t]
\centering
\caption{\textbf{Ablation study.} Following the Gen.\ Video Consistency setting, we evaluate the consistency among generated videos.}
\label{tab:ablation}
\small
\setlength{\tabcolsep}{5pt}
\renewcommand{\arraystretch}{\resulttablerowstretch}
\begin{tabular}{lccc}
\toprule
Configuration & PSNR$\uparrow$ & SSIM$\uparrow$ & LPIPS$\downarrow$ \\
\midrule
w/o TCR \& RCP & 21.09 & 0.766 & 0.216 \\
w/o RCP & 22.03 & 0.780 & 0.203 \\
w/o TCR & 22.21 & 0.788 & 0.196 \\
Full (Random) & 22.20 & 0.788 & 0.197 \\
Full (Strided) & 22.06 & 0.786 & 0.198 \\
Full (Sliding) & \textbf{22.63} & \textbf{0.796} & \textbf{0.191} \\
\bottomrule
\end{tabular}
\end{table}

\PAR{Implementation details.}
We build \sysname{} on top of Wan2.2-TI2V-5B~\cite{wan2025wan}, a 5B-parameter DiT-based video diffusion model whose high-compression VAE makes it particularly efficient for multi-view generation tasks.
Training is conducted at $704{\times}1280$ resolution with a learning rate of $1 \times 10^{-5}$ on 128 H20-3E GPUs, following the 3-stage curriculum described in Sec.~\ref{sec:training}. The three stages take approximately 0.5, 1, and 1.5 days, respectively.
At inference time, we use 20 denoising steps.
For Target Context Routing, we partition target views into four-view groups during target-view generation and set $t_s/T=0.2$.
See Supp.~\suppref{sec:supp_training}--\suppref{sec:supp_inference} for details.

\PAR{Evaluation protocol.}
For quantitative evaluation, we use 10 DNA-Rendering~\cite{cheng2023dna} and 3 DyMVHumans~\cite{zheng2024dymvhumans} test scenes unseen during training, selecting 16 approximately uniformly distributed cameras and 98 frames per scene.
Given a single front-view source video, \sysname{} generates 16 uniformly spaced target-view videos.
We report three complementary evaluation dimensions:
(i)~\emph{4DGS reconstruction}, where all 16 generated views are used to reconstruct a 4DGS model via FreeTimeGS~\cite{wang2025freetimegs}, and the rendered views are compared against ground-truth videos;
(ii)~\emph{generated video consistency}, where we hold out 4 evenly spaced views as the test set and use the remaining 12 views to train 4DGS, then compare the 4DGS renderings at test views against the corresponding generated videos to measure the 3D consistency among the generated views;
and (iii)~\emph{generated video reconstruction}, directly comparing all 16 generated videos against ground-truth videos.
All dimensions are evaluated using PSNR$\uparrow$, SSIM$\uparrow$, and LPIPS$\downarrow$.

\PAR{Baselines.}
We compare against three representative methods spanning different conditioning paradigms:
MV-Performer~\cite{zhi2025mvperformer}, an explicit geometry-conditioned human-specific novel view synthesis model natively trained on MVHumanNet++~\cite{li2025mvhumannetpp};
TrajectoryCrafter~\cite{yu2025trajectorycrafter}, an explicit geometry-conditioned camera control video model that uses depth-based warping;
and ReCamMaster$^\dagger$~\cite{bai2025recammaster}, an implicit camera-conditioned video model based on video conditioning.
$\dagger$ denotes our fine-tuned version trained with the same protocol and datasets and equipped with the same RCP module and TCR strategy for fair comparison, while MV-Performer and TrajectoryCrafter are evaluated zero-shot with their released weights.
All baselines are evaluated with the same source videos and target viewpoints.
The DNA-Rendering test sequences are held out from our training data, and DyMVHumans is out-of-distribution for all methods.

\subsection{Comparison with State-of-the-Art}
\label{sec:comparison}

\PAR{Quantitative results.}
Table~\ref{tab:comparison} presents the quantitative comparison on DNA-Rendering and DyMVHumans.
\sysname{} outperforms all baselines across all three evaluation dimensions.
For generated-video consistency, its leading scores indicate more coherent inputs for 4D lifting.
For generated-video reconstruction, its leading scores indicate more faithful novel views than the compared implicit-camera and dense-geometry baselines.
These consistency gains translate to the strongest 4DGS reconstruction results.

MV-Performer generates reasonable frontal views but distorts side and back views due to limited training diversity.
TrajectoryCrafter accumulates depth errors and often fails under front-to-back changes, while ReCamMaster's inaccurate camera control degrades 4DGS reconstruction despite plausible videos.

\PAR{Qualitative results.}
Fig.~\ref{fig:comparison} presents visual comparisons on the DNA-Rendering test set.
MV-Performer distorts body shape and pose at side and back views, and TrajectoryCrafter suffers severe geometry distortion and front-to-back failure.
ReCamMaster produces plausible videos, but imprecise camera control causes cross-view misalignment and noisy 4DGS renderings.
In contrast, \sysname{} remains geometrically accurate and detailed across viewpoints, including hallucination of unseen back-view content consistent with the front-view input.

\subsection{Ablation Study}
\label{sec:ablation}

We ablate the core components of \sysname{} on DNA-Rendering, selecting 8 challenging sequences, each with 16 uniformly distributed cameras and 121 frames. 
Following the Gen.\ Video Consistency setting in Sec.~\ref{sec:comparison}, we evaluate the relative consistency among generated videos. Results are reported in Table~\ref{tab:ablation}.

\PAR{Effect of Reference Context Packing.}
Without RCP, the model relies on monocular context alone and lacks appearance guidance from previously generated views, leading to inconsistent hallucinations in unseen viewpoints and degradation across all metrics.

\PAR{Effect of Target Context Routing.}
Without TCR, fixed view groups are denoised independently, causing structural and appearance inconsistencies across groups.
TCR rotates groupings at high noise to share context, improving all three metrics.
When RCP is additionally removed, all metrics deteriorate substantially, highlighting the complementary roles of RCP and TCR.

\PAR{Effect of routing strategy.}
With all components enabled, Random provides no measurable gain and Strided degrades all metrics relative to fixed grouping, whereas only Sliding improves all three, suggesting that preserving local view adjacency is important for routing.
See Supp.~\suppref{sec:supp_ablation} for details.

\PAR{Effect of 3D-aware skeleton conditioning.}
Compared with plain 2D rendering, depth buffering resolves front-back ambiguity for overlapping body parts (Fig.~\ref{fig:ablation}), improving geometric consistency.

\subsection{Generalization to In-the-Wild Videos}
\label{sec:wild_generalization}

\begin{figure*}[p]
\centering
\includegraphics[width=0.472\textwidth]{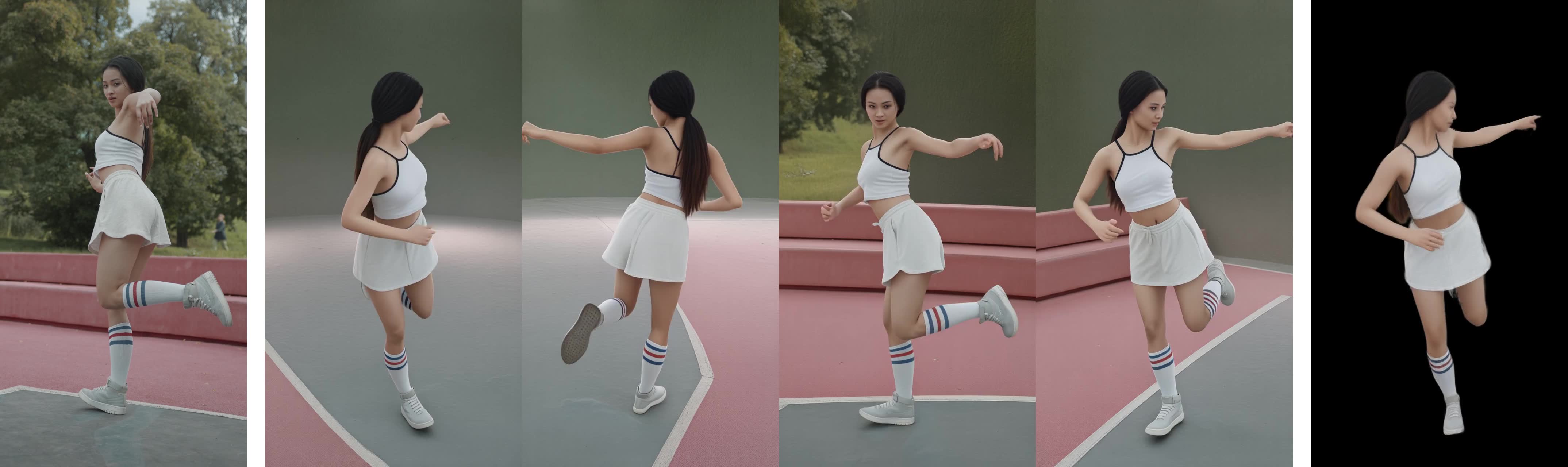}\hfill
\includegraphics[width=0.472\textwidth]{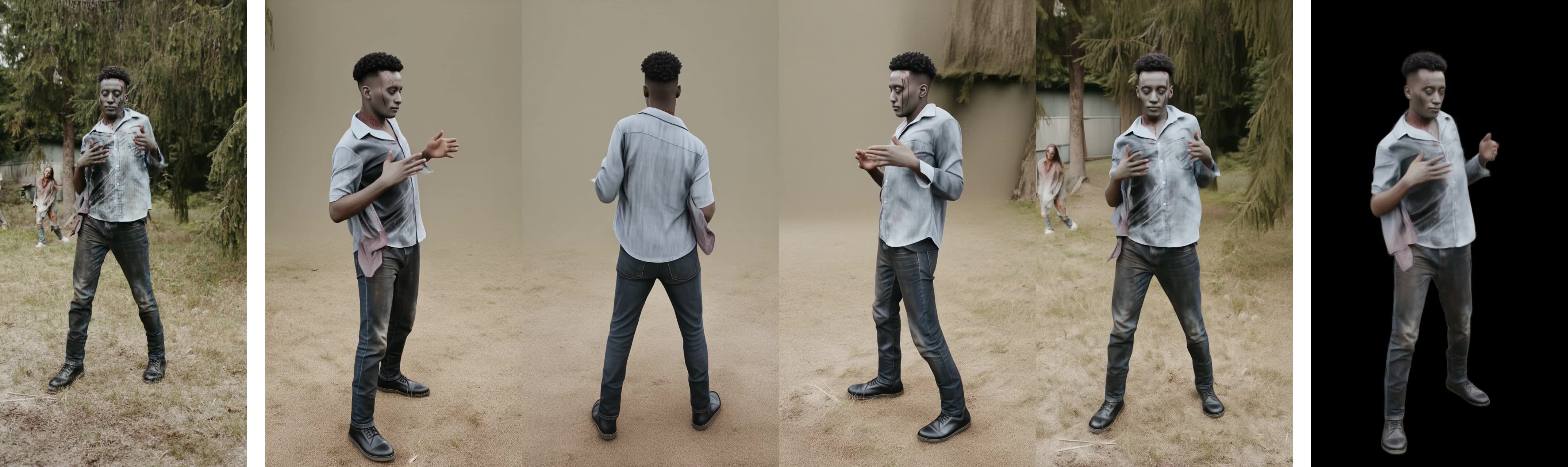}\\[1.5pt]
\includegraphics[width=0.472\textwidth]{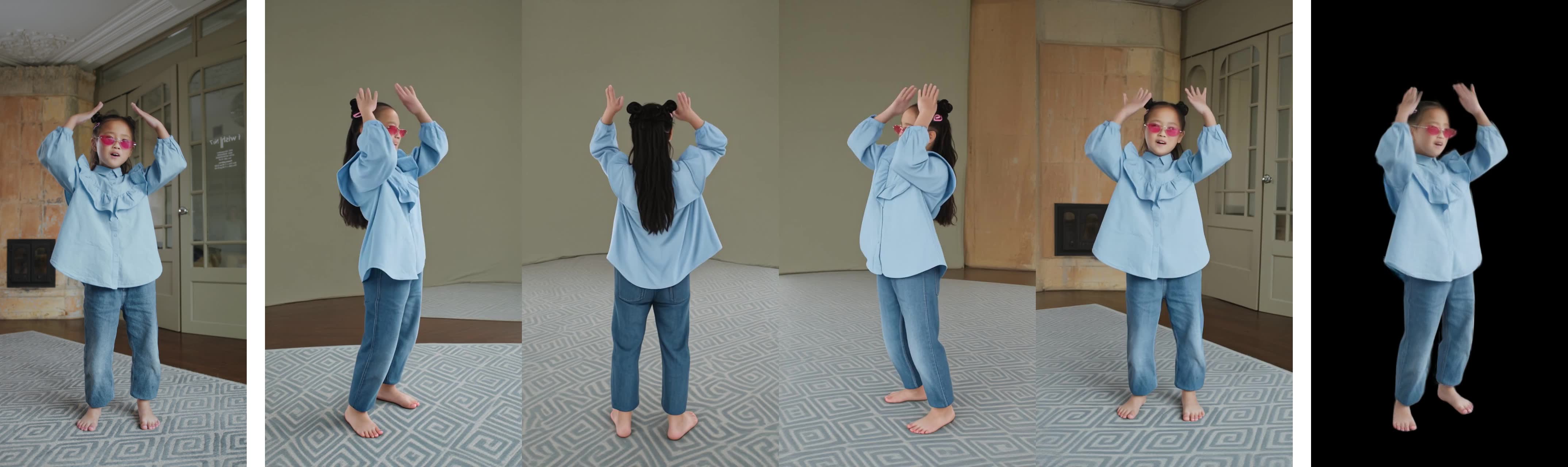}\hfill
\includegraphics[width=0.472\textwidth]{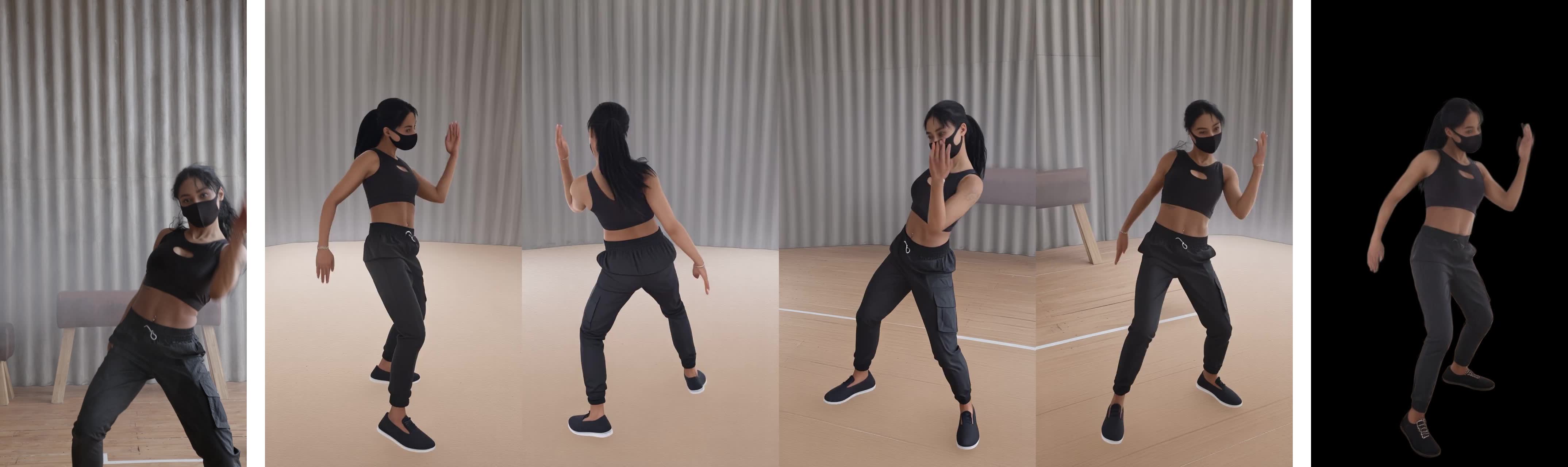}\\[1.5pt]
\includegraphics[width=0.472\textwidth]{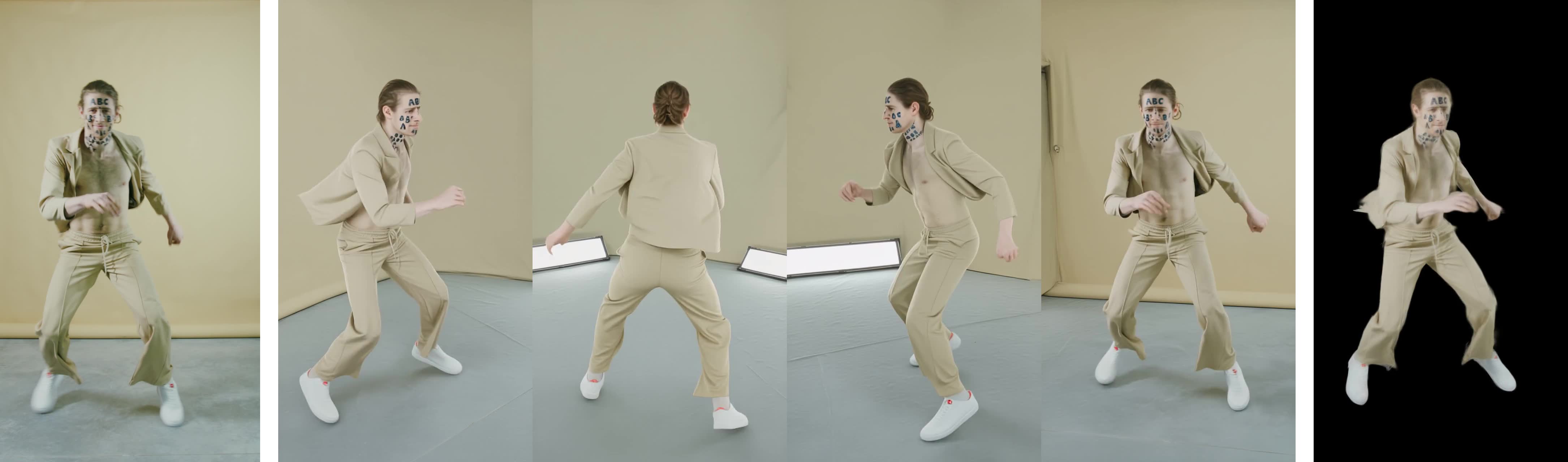}\hfill
\includegraphics[width=0.472\textwidth]{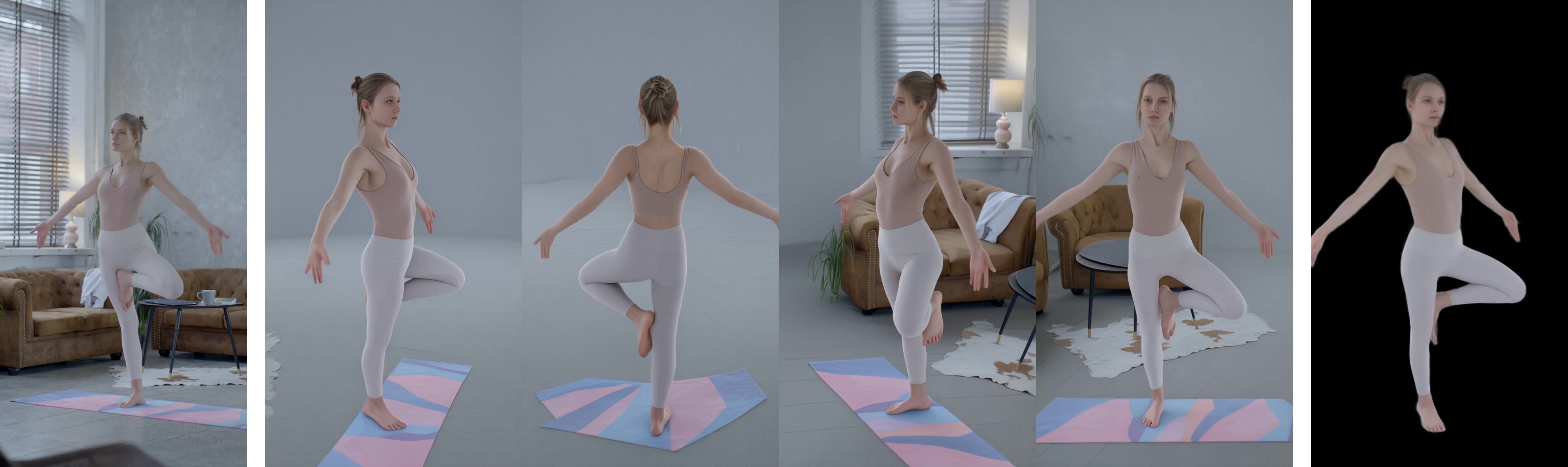}\\[1.5pt]
\includegraphics[width=0.472\textwidth]{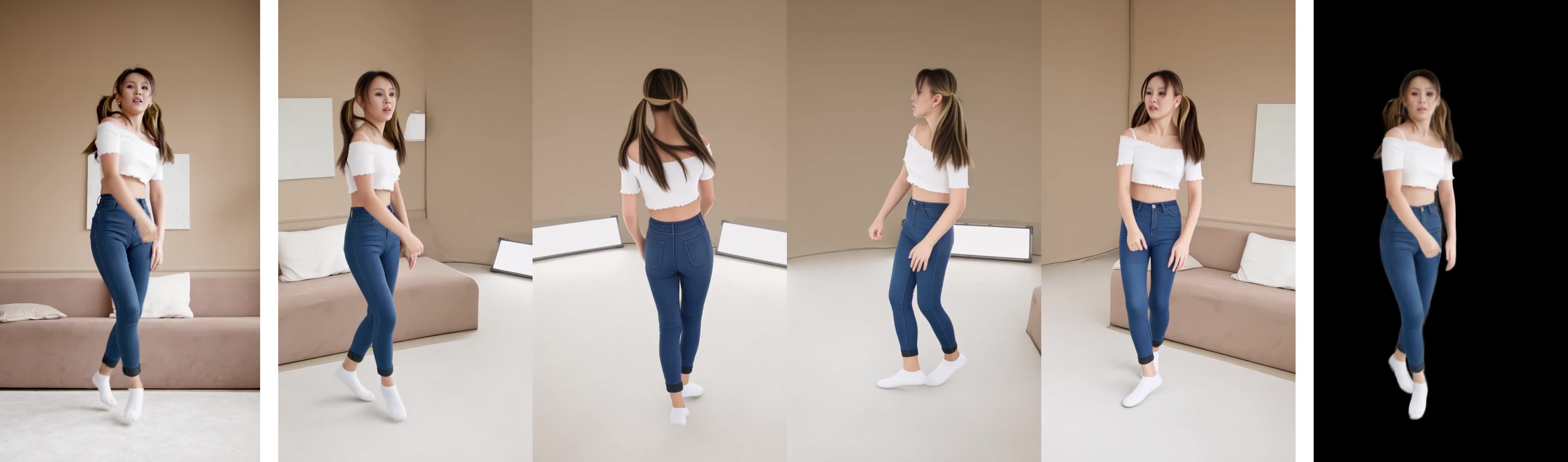}\hfill
\includegraphics[width=0.472\textwidth]{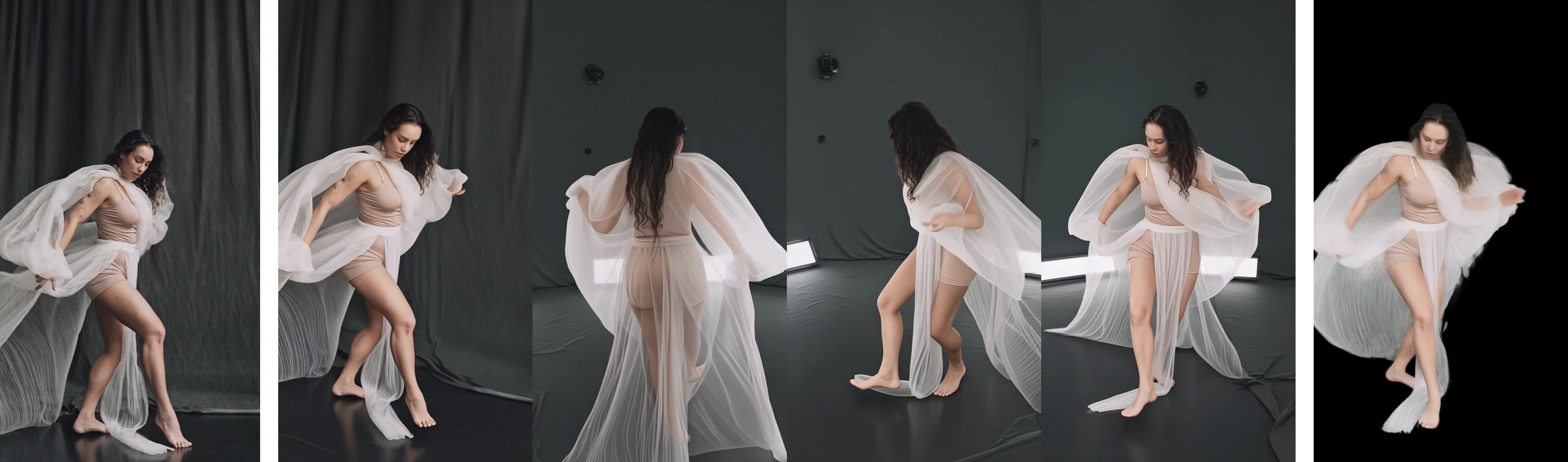}\\[1.5pt]
\includegraphics[width=0.472\textwidth]{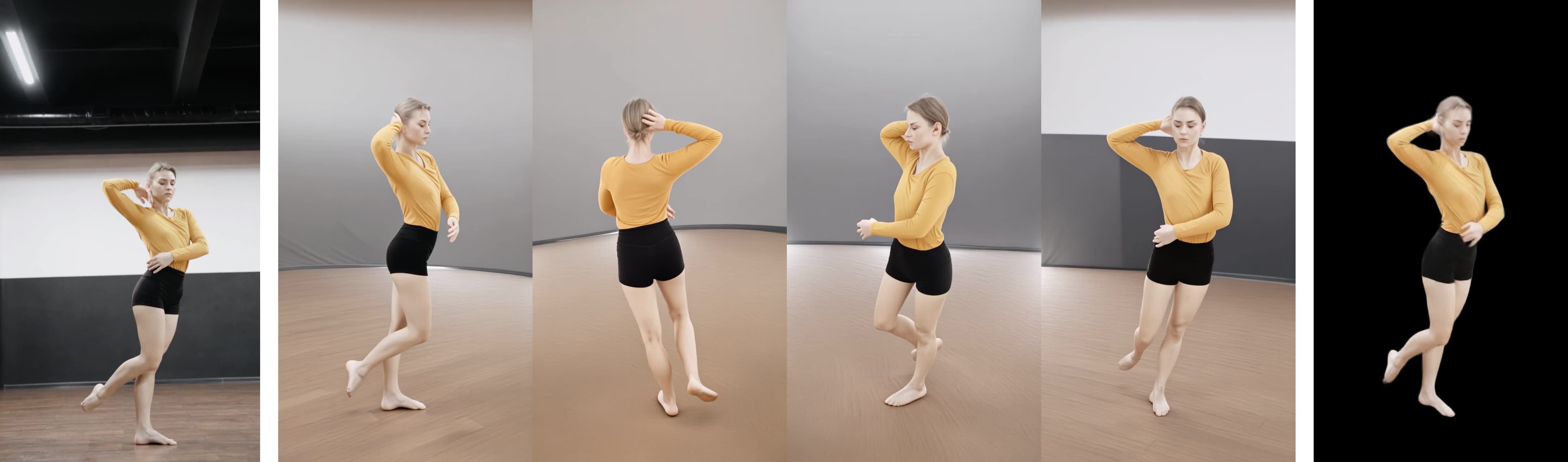}\hfill
\includegraphics[width=0.472\textwidth]{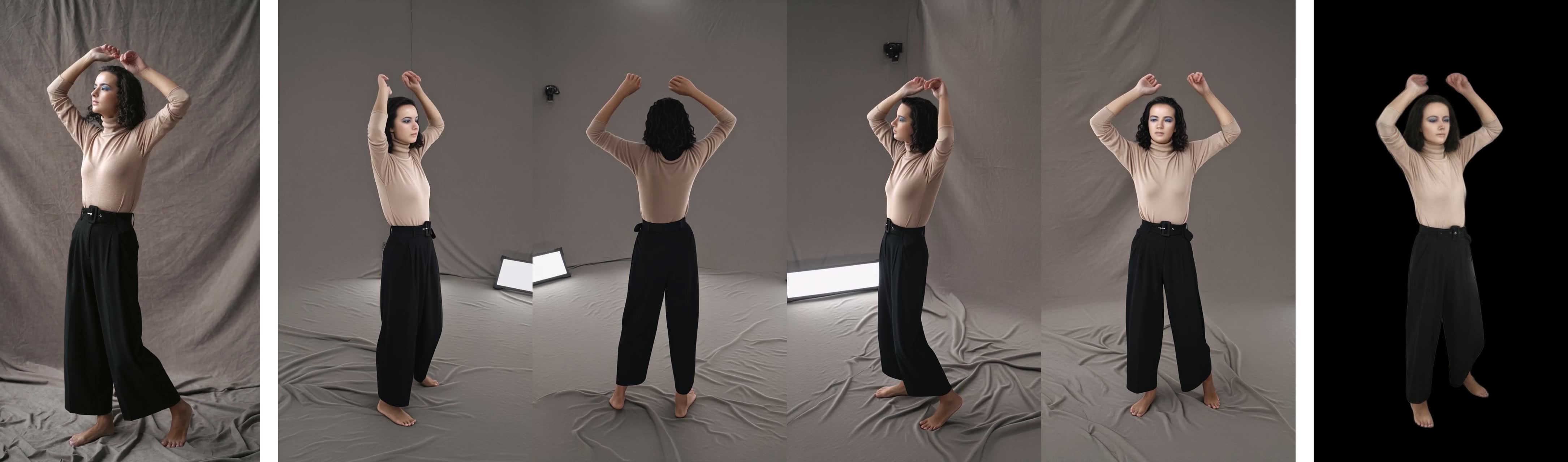}\\[1.5pt]
\includegraphics[width=0.472\textwidth]{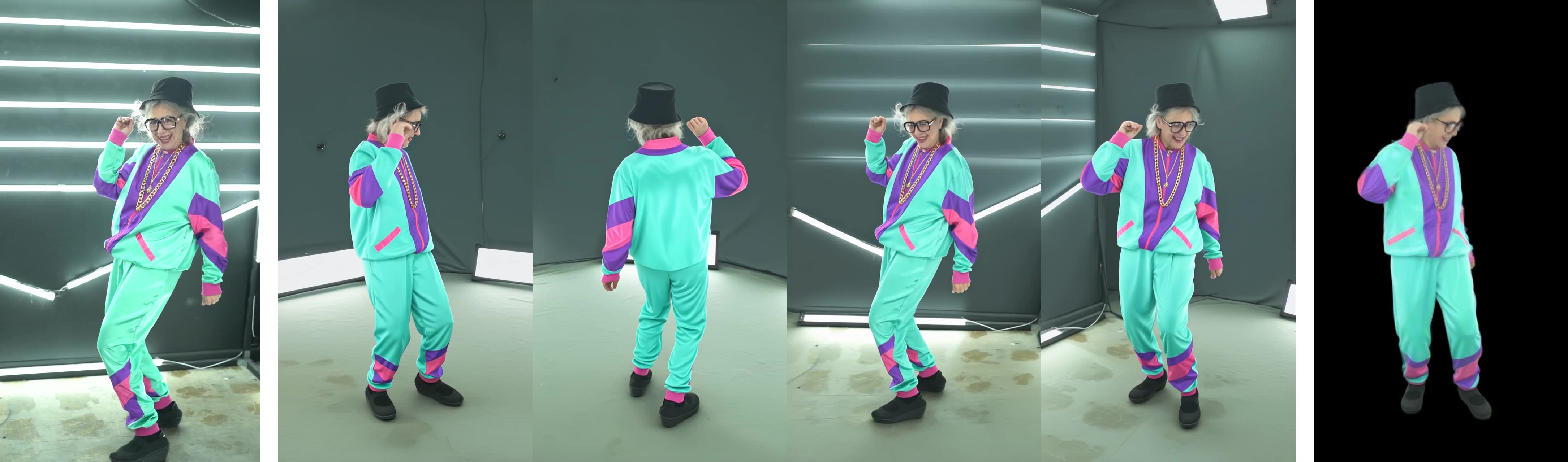}\hfill
\includegraphics[width=0.472\textwidth]{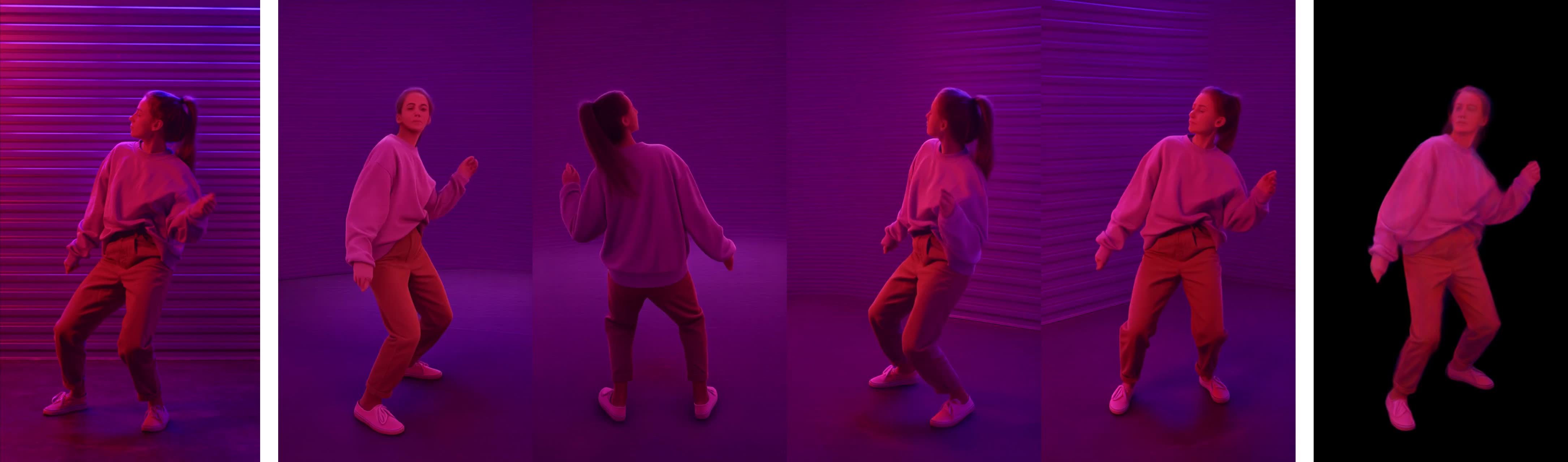}\\[1.5pt]
\includegraphics[width=0.472\textwidth]{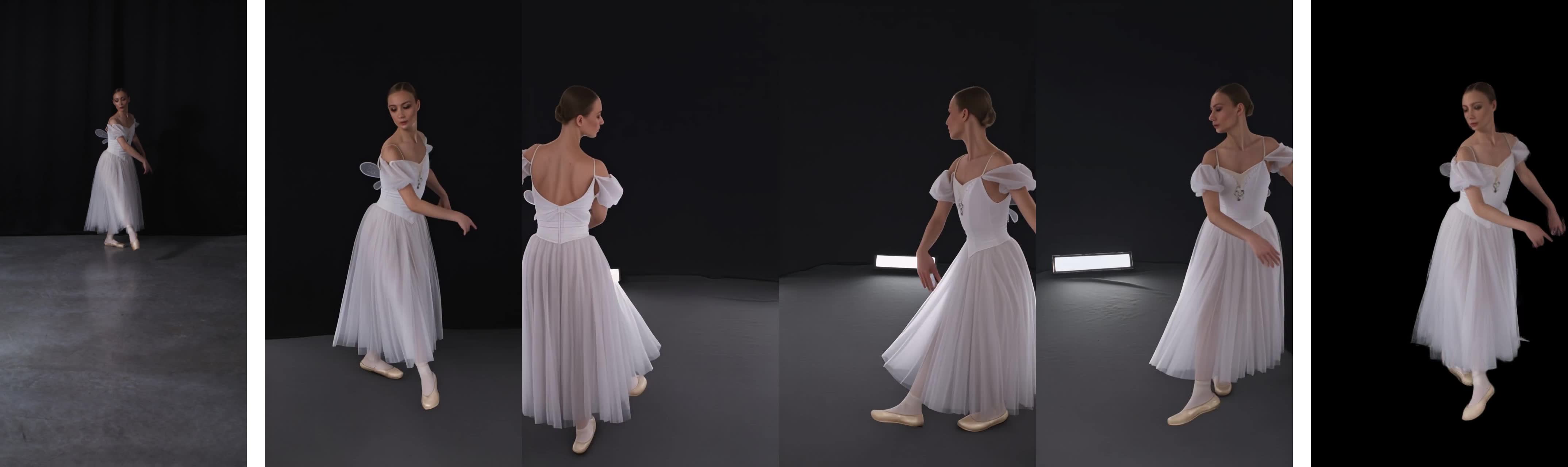}\hfill
\includegraphics[width=0.472\textwidth]{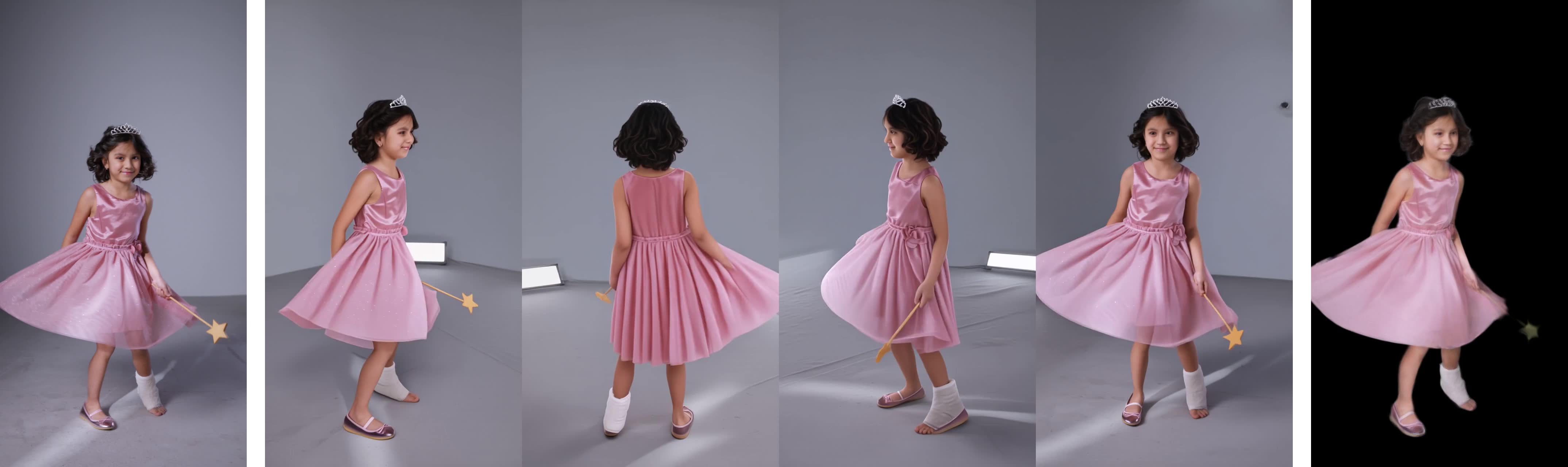}\\[1.5pt]
\includegraphics[width=0.472\textwidth]{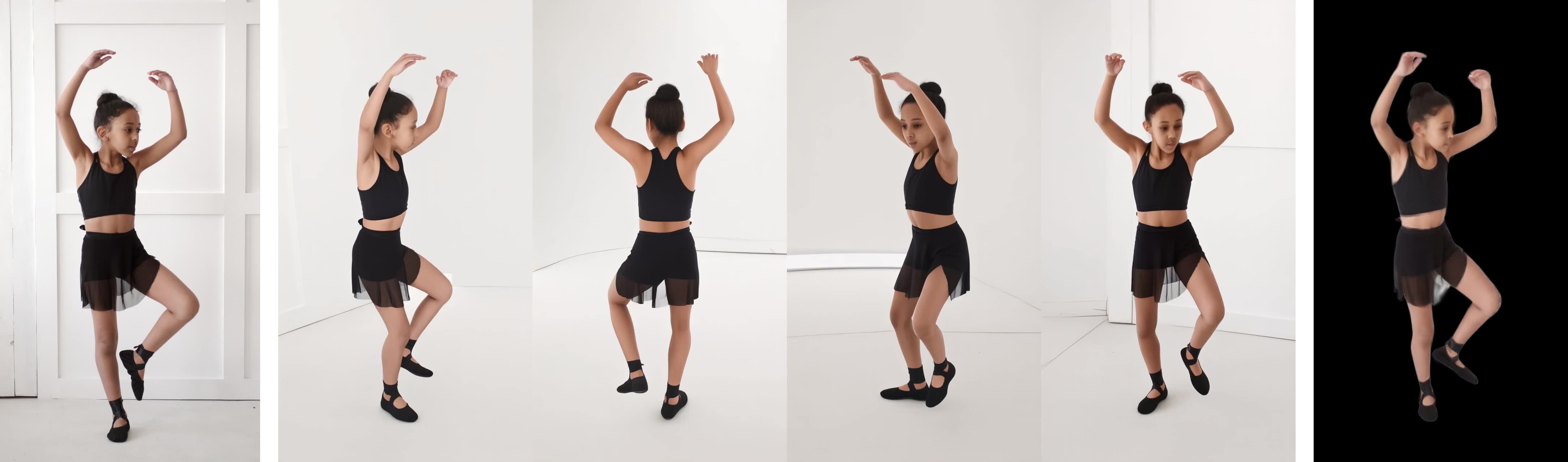}\hfill
\includegraphics[width=0.472\textwidth]{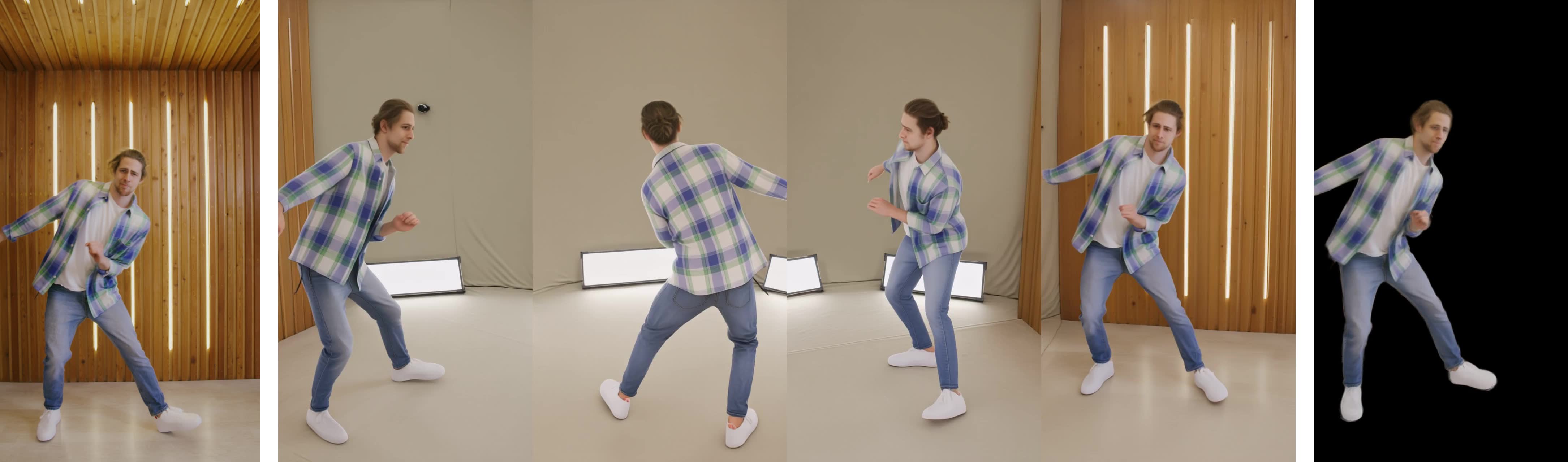}
\caption{\textbf{Robust generalization to diverse in-the-wild human videos.} For each example, we show the source video (left), generated target-view videos (middle, 4 of 16 views), and a 4DGS novel-view rendering (right). See the project page for dynamic results.}
\Description{Sixteen in-the-wild examples are arranged in two columns. Each example shows a source human video frame, four generated target viewpoints, and a novel-view rendering of the reconstructed 4D Gaussian avatar.}
\label{fig:wild_demo}
\end{figure*}

\PAR{In-the-Wild Results.}
We further evaluate \sysname{} on diverse in-the-wild human-centric videos.
As shown in Fig.~\ref{fig:wild_demo}, \sysname{} generalizes robustly beyond the controlled evaluation datasets, producing consistent target-view videos and high-quality 4DGS renderings across diverse human-centric videos.

\PAR{Challenging Cases.}
Fig.~\ref{fig:challenging} shows our tests on challenging inputs, including back-view source videos, complex subject appearance, and complex human motions.
\sysname{} remains robust in these settings, maintaining coherent target-view generation and high-quality 4DGS reconstruction.
Limitations and failure cases are analyzed in Supp.~\suppref{sec:supp_limitations}.

\section{Conclusion}
\label{sec:conclusion}

We have presented \sysname{}, a novel framework for reconstructing high-fidelity 4D humans from monocular videos.
By using 3D skeletons as sparse yet robust geometric conditioning, together with Reference Context Packing for scalable appearance guidance and Target Context Routing for reducing cross-group drift, \sysname{} generates reconstruction-grade multi-view videos that significantly outperform prior methods in both video quality and downstream 4DGS reconstruction.

\section*{Ethics and Impact}
\sysname{} synthesizes realistic human videos, which poses potential risks of deepfake misuse, identity privacy violation, and copyright infringement.
We firmly oppose any malicious use of our work, and advocate that generated results be clearly disclosed as synthetic and created only with the consent of the depicted subjects.

\begin{acks}
This work was partially supported by National Key R\&D Program of China (No.~2024YFB2809105), NSFC (No.~U24B20154), Zhejiang Provincial Natural Science Foundation of China (No.~LR25F020003), Ant Group, and Information Technology Center and State Key Lab of CAD\&CG, Zhejiang University.
\end{acks}

\clearpage
\bibliographystyle{ACM-Reference-Format}
\bibliography{main}

\clearpage
\begingroup
\setcounter{dbltopnumber}{3}
\setcounter{topnumber}{3}
\setcounter{totalnumber}{3}
\renewcommand{\dbltopfraction}{0.99}
\renewcommand{\floatpagefraction}{0.99}
\setlength{\dblfloatsep}{8pt}
\makeatletter
\setlength{\@dblfptop}{0pt}
\setlength{\@dblfpsep}{8pt}
\setlength{\@dblfpbot}{0pt plus 1fil}
\makeatother
\begin{figure*}[p]
\centering
\captionsetup{skip=0pt}
\includegraphics[width=\textwidth]{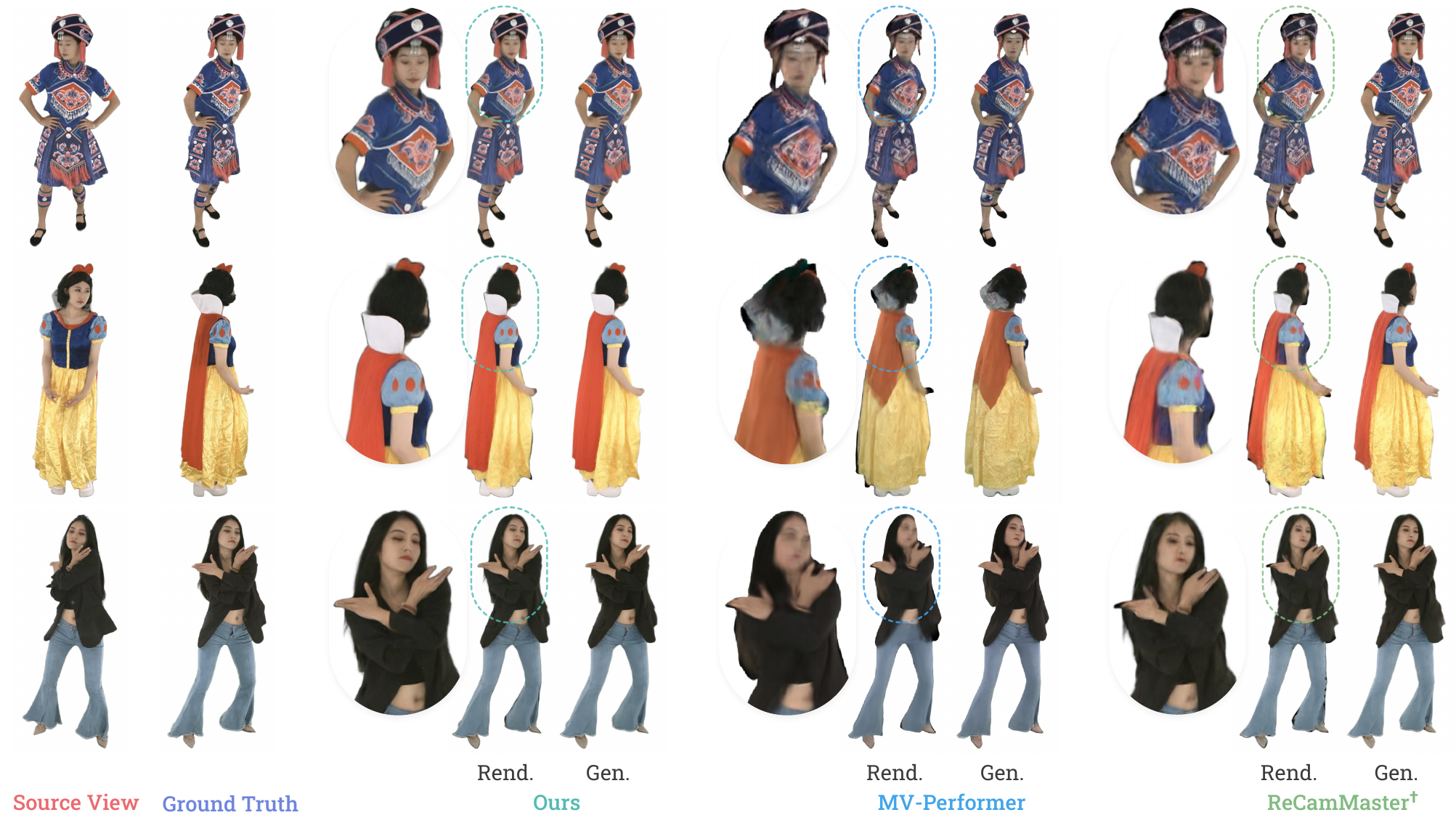}
\caption{\textbf{Qualitative comparison with baselines.} We show target-view generated videos (Gen.) and their corresponding 4DGS renderings (Rend.) across diverse human-centric videos. \sysname{} produces geometrically accurate and visually detailed results across viewpoints, while the baselines suffer from inaccurate camera control (our fine-tuned ReCamMaster$^\dagger$) or geometric distortions (MV-Performer). See the project page for dynamic results.}
\Description{A grid compares generated target views and reconstructed 4D renderings from several methods on diverse human videos. The proposed method preserves body geometry, clothing appearance, and camera viewpoint more consistently than ReCamMaster and MV-Performer.}
\label{fig:comparison}
\end{figure*}

\begin{figure*}[p]
\centering
\captionsetup{skip=0pt}
\includegraphics[width=\textwidth]{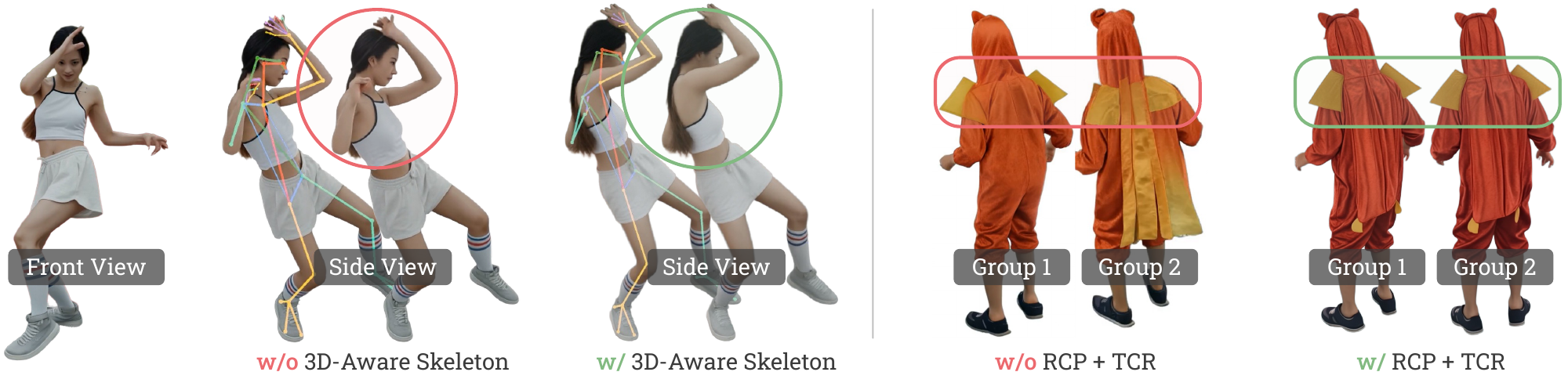}
\caption{\textbf{Qualitative ablation results.} \textbf{Left:} 3D-aware skeleton conditioning resolves the inherent ambiguity of 2D skeletons, guiding the model to generate geometrically correct content. \textbf{Right:} RCP and TCR maintain an effective cross-view context, leading to consistent appearance across generated views.}
\Description{Side-by-side ablation grids compare outputs with and without 3D-aware skeleton conditioning, Reference Context Packing, and Target Context Routing. The complete model has more accurate body geometry and more consistent appearance across views.}
\label{fig:ablation}
\end{figure*}

\begin{figure*}[p]
\centering
\captionsetup{skip=0pt}
\IfFileExists{figures/assets/challenging.pdf}{%
  \includegraphics[width=\textwidth]{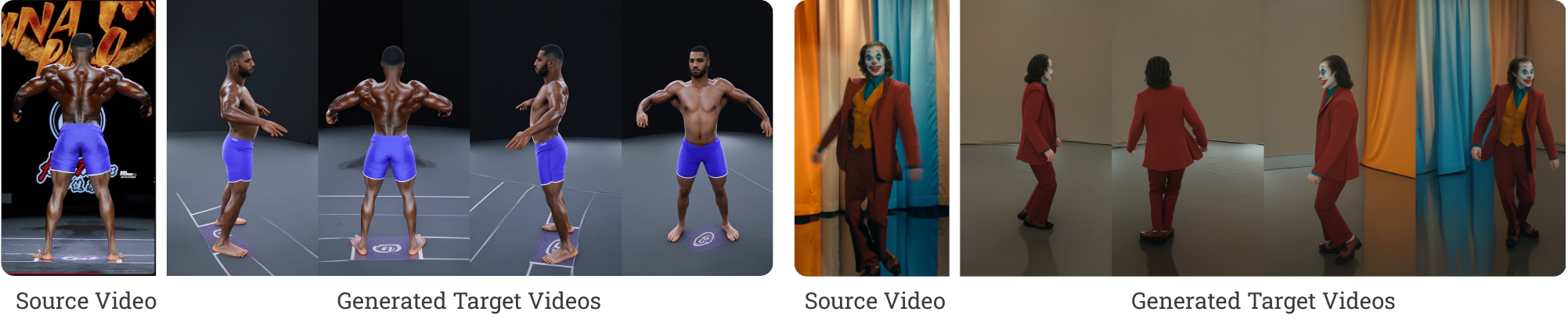}%
}{%
  \fbox{\rule{0pt}{0.52\textwidth}\rule{0.96\textwidth}{0pt}}%
}
\caption{\textbf{Challenging cases.} \sysname{} robustly handles back-view source videos (left), complex subject appearance (right), and complex human motions.}
\Description{Two challenging examples show source frames and generated viewpoints for a person initially seen from the back and a person with complex clothing and appearance.}
\label{fig:challenging}
\end{figure*}

\clearpage
\endgroup

\clearpage
\maketitlesupplementary{\papertitle}
\appendix
\section{Model Details}
\label{sec:supp_model}

\PAR{Multiview self-attention.}
The multiview self-attention layers share the same architecture and weights as the video (temporal) self-attention layers in the base Wan2.2 DiT~\cite{wan2025wan}, differing only in how tokens are arranged.
In multiview self-attention, we rearrange tokens to $(f, v{\cdot}h{\cdot}w, d)$, allowing tokens from different viewpoints at the same timestep to directly attend to each other.
These two attention modules jointly achieve 4D information exchange across all views and frames.
All parameters are initialized from the base model's temporal self-attention layers, so the pretrained temporal coherence serves as a natural starting point for learning cross-view consistency.

\PAR{Multi-scale patchify layers.}
The standard Wan2.2 patchify layer is a Conv3d with kernel/stride $(1,2,2)$.
The RCP $2{\times}$ and $4{\times}$ patchify layers use kernel/stride $(1,4,4)$ and $(1,8,8)$, producing $\frac{1}{4}$ and $\frac{1}{16}$ the tokens, respectively.
Following FramePack~\cite{zhang2025framepack}, we initialize them by tiling the pretrained $(1,2,2)$ kernel spatially and dividing by the area ratio ($4$ for $2{\times}$, $16$ for $4{\times}$) to preserve activation variance.

\PAR{3D-aware skeleton encoder.}
The skeleton encoder $g_\phi$ takes the depth-buffered RGB skeleton video as input and outputs a DiT-resolution residual added to the noisy latent tokens.
It consists of 10 Conv3d layers (5 strided, 5 non-strided, all with SiLU) followed by a $1{\times}1{\times}1$ final projection.
Channels progress as $3 \to 16 \to 32 \to 64 \to 128 \to 256 \to d$; strided convolutions use kernel $(3,4,4)$, yielding $32{\times}$ spatial and $4{\times}$ temporal downsampling.
The final projection is zero-initialized, so the residual is initially zero and the pretrained DiT behavior is preserved at the start of training.
We prepend 3 replicated copies of the first frame before feeding the sequence to the skeleton encoder to match the Wan2.2 VAE encoding pattern, which maps $4n{+}1$ input frames to $n{+}1$ latent frames.

\section{Dataset Details}
\label{sec:supp_dataset}

\begin{figure*}[t]
\centering
\includegraphics[width=\textwidth]{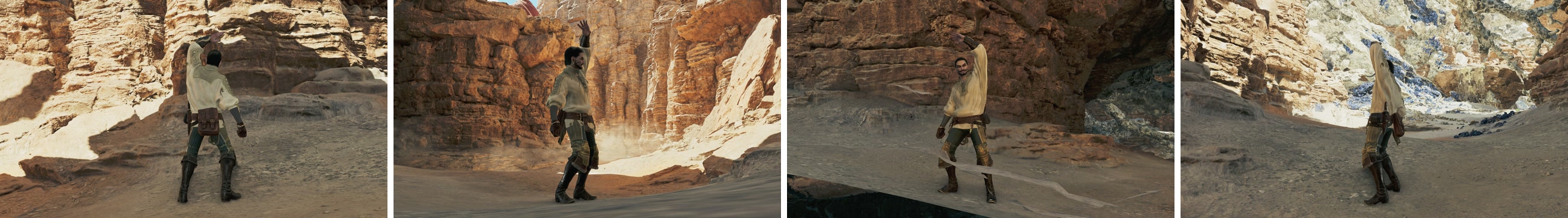}\\[1pt]
\includegraphics[width=\textwidth]{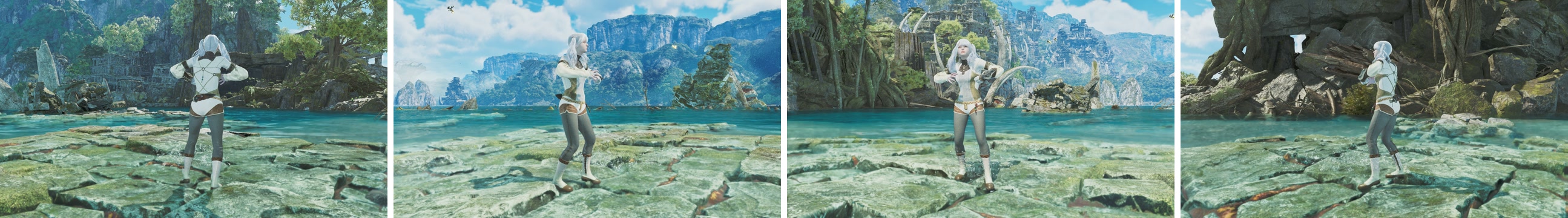}\\[1pt]
\includegraphics[width=\textwidth]{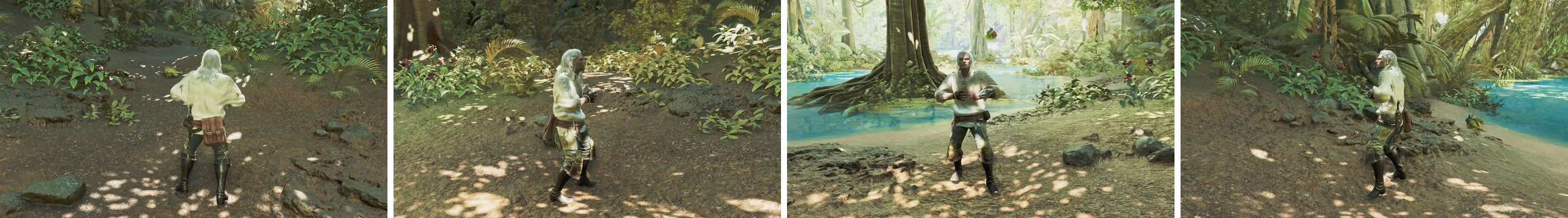}\\[1pt]
\includegraphics[width=\textwidth]{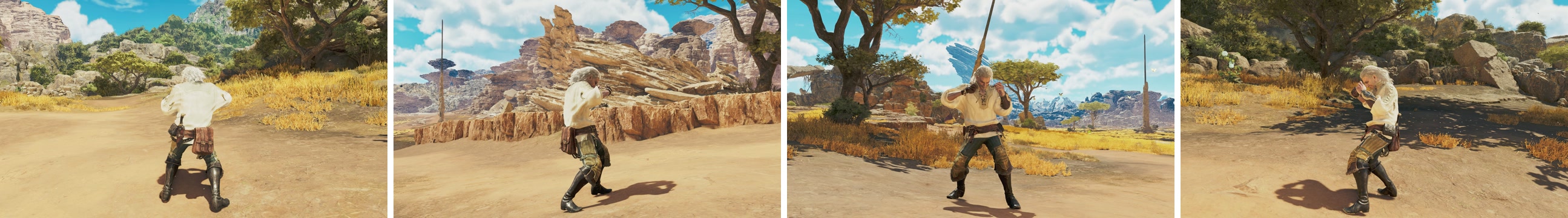}\\[1pt]
\includegraphics[width=\textwidth]{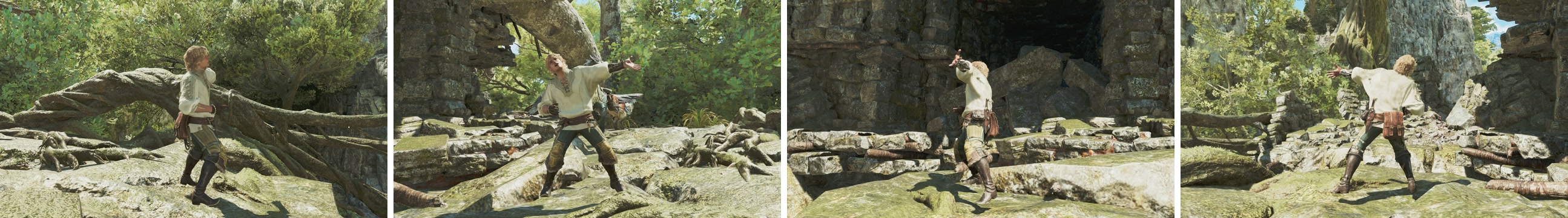}\\[1pt]
\includegraphics[width=\textwidth]{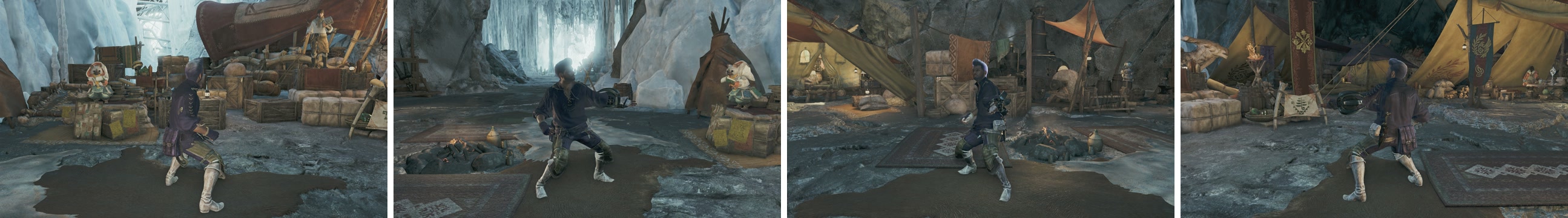}\\[1pt]
\includegraphics[width=\textwidth]{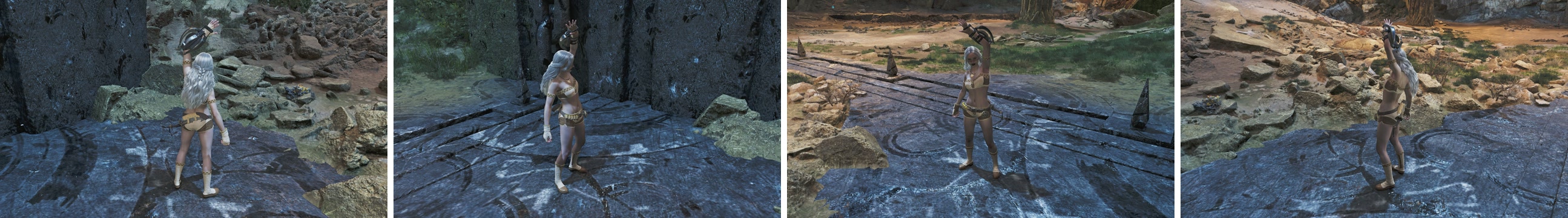}\\[1pt]
\includegraphics[width=\textwidth]{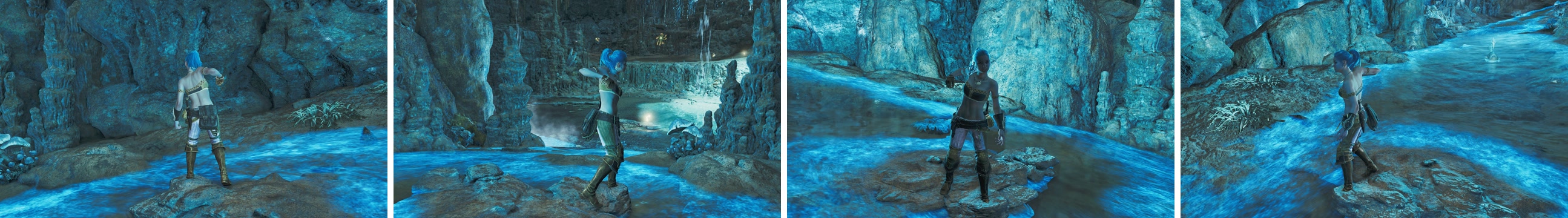}
\caption{\textbf{Representative MVGameHuman samples.} Each row shows one frame from four evenly spaced cameras in a synchronized 24-camera sequence captured with our in-house game data engine. MVGameHuman provides diverse actors, clothing, motions, lighting conditions, virtual scenes, and backgrounds for training multi-view human video generation.}
\Description{Eight rows of synthetic human scenes, each viewed simultaneously from four evenly spaced virtual cameras. The samples vary in actor, clothing, pose, lighting, environment, and background.}
\label{fig:supp_mvgamehuman}
\end{figure*}

\PAR{MVGameHuman.}
MVGameHuman is collected with our in-house game data engine.
It contains 38k synchronized multi-view human videos rendered at $2560{\times}1440$ resolution, covering 318 actors captured by 24 virtual cameras per sequence.
Fig.~\ref{fig:supp_mvgamehuman} visualizes representative samples from MVGameHuman.
Each row shows the same frame observed by four uniformly spaced cameras selected from the 24-camera rig.
The examples illustrate the dataset's synchronized multi-view coverage and variations in actors, clothing, motion, lighting, virtual scenes, and backgrounds.

\section{Training Details}
\label{sec:supp_training}

\PAR{Training setup.}
All stages fine-tune Wan2.2-TI2V-5B~\cite{wan2025wan} at $704{\times}1280$ resolution with a learning rate of $1 \times 10^{-5}$ and an LPIPS~\cite{zhang2018lpips} weight of $\lambda=0.25$.
The three stages take approximately 0.5, 1, and 1.5 days on 128 H20-3E GPUs; training converges stably with 32 or more GPUs, and we use 128 to accelerate it.

\PAR{Body-part-aware LPIPS cropping.}
Full-resolution LPIPS is memory intensive for long clips, so we use a body-part-aware sampler that prioritizes human regions, especially the face and hands, while retaining global coverage.
From target-view 2D keypoints, we derive clip-level boxes for the full body, face, and hands and decode one spatial crop per clip.
We sample the full body, face, left hand, and right hand with probabilities $0.2$, $0.2$, $0.1$, and $0.1$; the remaining $0.4$ is uniform.
Face and hand crops are box-centered, whereas full-body crops center on a point sampled within the box; missing boxes fall back to uniform sampling.
Each $256{\times}256$ crop is shared across all clip frames.

\PAR{Data sampling configuration.}
Tab.~\ref{tab:supp_training_sampling} summarizes the training sampling configuration.
For each iteration, a dataset and camera configuration are randomly selected according to the specified sampling weights.
To maintain roughly constant computational cost across configurations, the product of target cameras and frame count is kept approximately constant (\eg $6 \times 41 \approx 4 \times 61 \approx 1 \times 121$), so that the total number of tokens per forward-backward pass remains similar.
Source cameras are uniformly sampled from the listed options to train the RCP module's multi-scale patchify layers across different compression ratios: 1 source view uses $1{\times}$ patchify, 4 source views use $2{\times}$ patchify, and 8 source views combine $1{\times}$, $2{\times}$, and $4{\times}$ patchify layers.

\begin{table}[t]
\centering
\caption{\textbf{Training data sampling.} Source cameras are uniformly sampled from the listed options.}
\label{tab:supp_training_sampling}
\small
\setlength{\tabcolsep}{0pt}
\begin{tabular*}{\columnwidth}{@{\extracolsep{\fill}}lcccc@{}}
\toprule
Dataset & No.\ Tgt Cam & No.\ Src Cam & No.\ Frame & Weight \\
\midrule
MVGameHuman & 6 & 1\,/\,4\,/\,8 & 41 & 0.4 \\
MVGameHuman & 4 & 1\,/\,4\,/\,8 & 61 & 0.4 \\
MVGameHuman & 1 & 1\,/\,4\,/\,8 & 121 & 0.2 \\
SynCamVideo & 4 & 1\,/\,4 & 61 & 0.8 \\
SynCamVideo & 1 & 1\,/\,4\,/\,8 & 81 & 0.2 \\
DNA-Rendering & 6 & 1\,/\,4\,/\,8 & 41 & 0.4 \\
DNA-Rendering & 4 & 1\,/\,4\,/\,8 & 61 & 0.4 \\
DNA-Rendering & 1 & 1\,/\,4\,/\,8 & 121 & 0.2 \\
Pexels & 1 & 1 & 121 & 1.0 \\
TedTalk & 1 & 1 & 121 & 1.0 \\
\bottomrule
\end{tabular*}
\end{table}

\PAR{Stage-specific settings.}
Tab.~\ref{tab:supp_stage_settings} details the per-stage training configuration.
In Stage~1, we train exclusively on foreground-only DNA-Rendering~\cite{cheng2023dna} videos.
A 20\% probability of independently sampling the source frame range decouples pose from appearance: the model learns to follow the skeleton's pose while referencing the source video's appearance from a potentially different temporal window.
Stage~2 removes background masking and adds MVGameHuman and SynCamVideo to improve diversity and enable background modeling.
In Stage~3, we further include monocular datasets (Pexels and TedTalk) and drop finger keypoints from the skeleton input, since monocular finger keypoint detections are noisy; the model instead learns to infer hand details from the source video reference.

\begin{table}[t]
\centering
\caption{\textbf{Stage-specific training settings.} ``Indep.\ Src Prob'' denotes the probability of independently sampling the source frame range. Skeleton: B=body, H=hands, F=feet, Fi=fingers.}
\label{tab:supp_stage_settings}
\small
\setlength{\tabcolsep}{0pt}
\begin{tabular*}{\columnwidth}{@{\extracolsep{\fill}}clccl@{}}
\toprule
Stage & Dataset & Bg.\ Removal & Indep.\ Src Prob & Skeleton \\
\midrule
1 & DNA-Rendering & $\checkmark$ & 0.2 & B+H+F+Fi \\
2 & +\,MVGameHuman, SynCam. & -- & 0.0 & B+H+F+Fi \\
3 & +\,Pexels, TedTalk & -- & 0.0 & B+H+F \\
\bottomrule
\end{tabular*}
\end{table}

\section{HMR Details}
\label{sec:supp_hmr}

\PAR{Human motion recovery.}
Given a monocular video, we run GVHMR~\cite{shen2024gvhmr} to estimate a ground-aligned SMPL-X~\cite{pavlakos2019expressive} mesh sequence, then apply a sparse vertex-to-keypoint regressor to extract 70 3D keypoints of the Goliath vocabulary~\cite{khirodkar2026sapiens2}.
For skeleton rendering, we keep the body, foot, and palm-level hand keypoints (the wrist and five knuckles per hand), excluding face and individual finger joints.
Skeletons are rendered with body-part-specific coloring to help the model distinguish different body regions; each keypoint's camera-space depth drives the pixelwise z-buffer, which depends only on relative depth ordering and is thus invariant to absolute scale and shift.

\PAR{SMPL-X-to-Goliath70 regressor.}
The regressor predicts each of the 70 Goliath keypoints as a convex combination of a fixed set of nearby SMPL-X vertices, reducing keypoint extraction to a sparse weighted sum over mesh vertices.
To train it, we run SAM 3D Body~\cite{yang2026sam3dbody} on DNA-Rendering~\cite{cheng2023dna} frames to obtain Momentum Human Rig (MHR) parameters, which yield both the target Goliath keypoints and, through the official MHR-to-SMPL-X conversion, the paired SMPL-X vertices; frames with conversion error above 1\,mm are discarded.
Support vertices are selected as each keypoint's nearest SMPL-X vertices and kept fixed, and the convex weights are optimized with a Smooth-L1 loss.
On held-out scenes, the regressor achieves a 3.5\,mm mean keypoint error, compared with 14.0\,mm for a nearest-vertex baseline.

\section{Inference Details}
\label{sec:supp_inference}

Taking the multi-view skeleton conditions prepared in Sec.~\ref{sec:supp_hmr} as input, the inference pipeline generates all target-view videos and reconstructs the final 4DGS model as detailed below.

\PAR{Multi-GPU inference.}
Target Context Routing supports both single-GPU sequential execution and multi-GPU parallel denoising for faster inference.
Tab.~\ref{tab:supp_multigpu} lists three commonly used configurations.
A single-layer setup with 16 cameras suffices for free-viewpoint rendering at limited pitch angles.
Two layers with 32 cameras cover most scenarios, while three layers with 48 cameras accommodate subjects with complex clothing or extreme motions that require denser view coverage.

\begin{table}[t]
\centering
\setlength{\tabcolsep}{4pt}
\caption{\textbf{Multi-GPU inference configurations.} Configurations are shown for different camera setups.}
\label{tab:supp_multigpu}
\small
\begin{tabular}{cccc}
\toprule
No.\ Layers & No.\ Cam\,/\,Layer & No.\ GPUs & No.\ Tgt Cam\,/\,GPU \\
\midrule
1 & 16 & 4 & 4 \\
2 & 16 & 8 & 4 \\
3 & 16 & 8 & 6 \\
\bottomrule
\end{tabular}
\end{table}

\PAR{4DGS reconstruction.}
We employ FreeTimeGS~\cite{wang2025freetimegs}, an enhanced version of 4DGS~\cite{wu20244dgs} and LongVolcap~\cite{xu2024longvolcap}, to reconstruct 4D human performances from the generated multi-view videos.
The 4D Gaussian primitives are initialized with coarse geometry obtained via space carving from the predicted foreground masks.
We optimize the model using the Adam optimizer with a learning rate of $1.6 \times 10^{-4}$ for 50k iterations on sequences of 16 cameras and 121 frames.

\PAR{Inference efficiency.}
The full \sysname{} pipeline for 4D human reconstruction involves three stages with the following approximate timings:
\begin{enumerate}[nosep,leftmargin=*]
    \item \textbf{Preprocessing} (Sec.~\ref{sec:supp_hmr}): GVHMR inference and multi-view skeleton rendering, taking ${\sim}$2\,min on a single RTX 4090.
    \item \textbf{Multi-view video generation}: generating 4 videos of 121 frames each with 20 denoising steps, taking ${\sim}$7\,min on a single H20 GPU. Empirically, the strong conditioning guidance of our model allows reducing to 10 denoising steps with minimal quality degradation.
    \item \textbf{4DGS training}: FreeTimeGS optimization from the generated multi-view videos, taking ${\sim}$30\,min on a single RTX 4090.
\end{enumerate}

\section{Evaluation Details}
\label{sec:supp_eval}

We provide additional details on the evaluation setup for each baseline and our method.
All methods are evaluated on the same test sequences from DNA-Rendering~\cite{cheng2023dna} (10 scenes) and DyMVHumans~\cite{zheng2024dymvhumans} (3 scenes), with 16 approximately uniformly distributed cameras and 98 frames per scene.
For DNA-Rendering, we use camera 22 as the source, cameras [01, 04, $\ldots$, 46] (stride 3) as targets, [04, 16, 28, 40] as RCP references, and [01, 13, 25, 37] as held-out views for consistency evaluation.
DyMVHumans uses a 60-camera array. We use camera 29 as the source, cameras [02, 04, 06, 09, 14, 19, 24, 29, 33, 38, 43, 48, 52, 54, 57, 59] as targets, [04, 19, 38, 54] as RCP references, and [02, 14, 33, 52] as held-out views for consistency evaluation.
For each dataset, the remaining 12 target views are used to optimize 4DGS for consistency evaluation.
The specific test sequences are listed below:
\begin{itemize}[nosep,leftmargin=*]
    \item \textbf{DNA-Rendering}:\\
    0012\_09, 0019\_06, 0025\_11, 0034\_04, 0094\_02,\\
    0124\_03, 0152\_01, 0165\_08, 0188\_02, 0219\_07.
    \item \textbf{DyMVHumans}:\\
    1080\_Dance\_Dunhuang\_Single\_f14,\\
    1080\_Sport\_Badminton\_Single\_f11,\\
    1080\_Sport\_Football\_Single\_m11.
\end{itemize}

\PAR{MV-Performer.}
MV-Performer is natively trained on MVHumanNet++~\cite{li2025mvhumannetpp}; we evaluate its officially released model zero-shot on both benchmarks.
Following the official MV-Performer pipeline, we construct the target-view condition from the source-view depth map.
We first estimate depth with Depth-Anything-3~\cite{lin2025da3} using the ground-truth camera parameters of the evaluation capture, so the prediction is already expressed in the dataset camera scale.
We run Depth-Anything-3 on all available views with ground-truth camera parameters to obtain reconstruction-grade depth, maximizing the generation accuracy of MV-Performer.
We then take the depth map of the source camera, warp it to each of the 16 target views, and convert the warped depth into the normal-map condition used by MV-Performer.
For inference, we split the 16 target cameras into two evenly spaced batches and process the 98-frame sequence in two 49-frame clips.
We keep the inference resolution at $832\times 480$, matching the official MV-Performer examples.

\PAR{TrajectoryCrafter.}
TrajectoryCrafter is trained on a hybrid of web-scale monocular videos and static multi-view data; we evaluate its officially released model zero-shot on both benchmarks.
TrajectoryCrafter uses warped source-video depth as the target-view condition.
We first predict source-video depth with DepthCrafter~\cite{hu2025depthcrafter} and align it to the dataset camera scale before warping.
We use a camera-aligned Depth-Anything-3~\cite{lin2025da3} depth map as reference and compute a per-sequence scale and shift from foreground pixels only.
The aligned depth is then warped to each of the 16 target views and fed to TrajectoryCrafter as the geometric condition.
To match the model's training setup, we additionally center-crop and resize the input frames to $672\times 384$ before all other processing steps.
Each inference run generates one target-view video of 49 frames, so the full 98-frame sequence is processed in two clips per target view.

\PAR{ReCamMaster.}
We fine-tune ReCamMaster~\cite{bai2025recammaster} on the same training datasets and with the same training settings as our model.
Specifically, we equip ReCamMaster with the same RCP module and TCR strategy used in \sysname{}, providing a controlled comparison between implicit camera-parameter conditioning and our explicit skeleton-geometry conditioning.

\PAR{\sysname{}.}
We first generate 4 uniformly spaced reference views from the source video in a single round.
We then generate the 16 target views in four-view groups with TCR, using the fixed RCP context built from the source video and these references.
Our model inherits the 121-frame generation length from Wan2.2-TI2V-5B~\cite{wan2025wan}; we truncate to the first 98 frames to match the test sequence length.

\subsection{Ablation Details}
\label{sec:supp_ablation}

We conduct the ablation study on eight DNA-Rendering scenes [0007\_04, 0166\_04, 0173\_02, 0592\_01, 0623\_01, 0700\_06, 0718\_06, 0811\_06].
We follow the DNA-Rendering comparison split above, omitting the reference views only for w/o RCP.

We order the 16 target views by azimuth and partition them into four-view groups.
At each dynamic step $n$, \emph{Sliding} applies a cumulative one-position circular shift, \emph{Random} applies a deterministic permutation with seed $42+n$, and \emph{Strided} reorders the views with stride $4-(n \bmod 4)$ before grouping.
All three variants use dynamic routing for the first 16 of 20 denoising steps and fixed contiguous groups for the final four steps ($t_s/T=0.2$), while w/o TCR uses fixed contiguous groups for all 20 steps.

\PAR{TCR switching-time sweep.}
We vary only $t_s/T$ while keeping all other ablation settings fixed.
With 20 denoising steps, $t_s/T=1$ denotes fixed contiguous grouping throughout, while $t_s/T=0$ denotes sliding throughout; intermediate values use sliding first and fixed grouping for the final $20t_s/T$ steps.
As shown in Tab.~\ref{tab:supp_tcr_sweep}, increasing the number of sliding steps improves consistency until $t_s/T=0.2$.
Further decreasing $t_s/T$ yields no measurable consistency gain and can even slightly degrade individual metrics; these non-monotonic fluctuations are comparable to variation from 4DGS optimization.
We therefore use $t_s/T=0.2$, the start of this saturation regime, as the default.

\begin{table}[t]
\centering
\caption{\textbf{TCR switching-time sweep.} Gen.\ Video Consistency when varying the number of sliding denoising steps.}
\label{tab:supp_tcr_sweep}
\small
\setlength{\tabcolsep}{4pt}
\begin{tabular}{ccccc}
\toprule
$t_s/T$ & Sliding steps & PSNR$\uparrow$ & SSIM$\uparrow$ & LPIPS$\downarrow$ \\
\midrule
1.00 & 0  & 22.2079 & 0.7880 & 0.1964 \\
0.75 & 5  & 22.3030 & 0.7898 & 0.1947 \\
0.50 & 10 & 22.4575 & 0.7925 & 0.1933 \\
0.25 & 15 & 22.6093 & 0.7955 & 0.1912 \\
0.20 & 16 & 22.6294 & 0.7963 & 0.1906 \\
0.15 & 17 & 22.6023 & 0.7961 & 0.1908 \\
0.10 & 18 & 22.6217 & 0.7967 & 0.1905 \\
0.05 & 19 & 22.6367 & 0.7972 & 0.1903 \\
0.00 & 20 & 22.6414 & 0.7971 & 0.1903 \\
\bottomrule
\end{tabular}
\end{table}

\section{Additional Results}
\label{sec:supp_results}

\begin{figure*}[t]
\centering
\includegraphics[width=\textwidth]{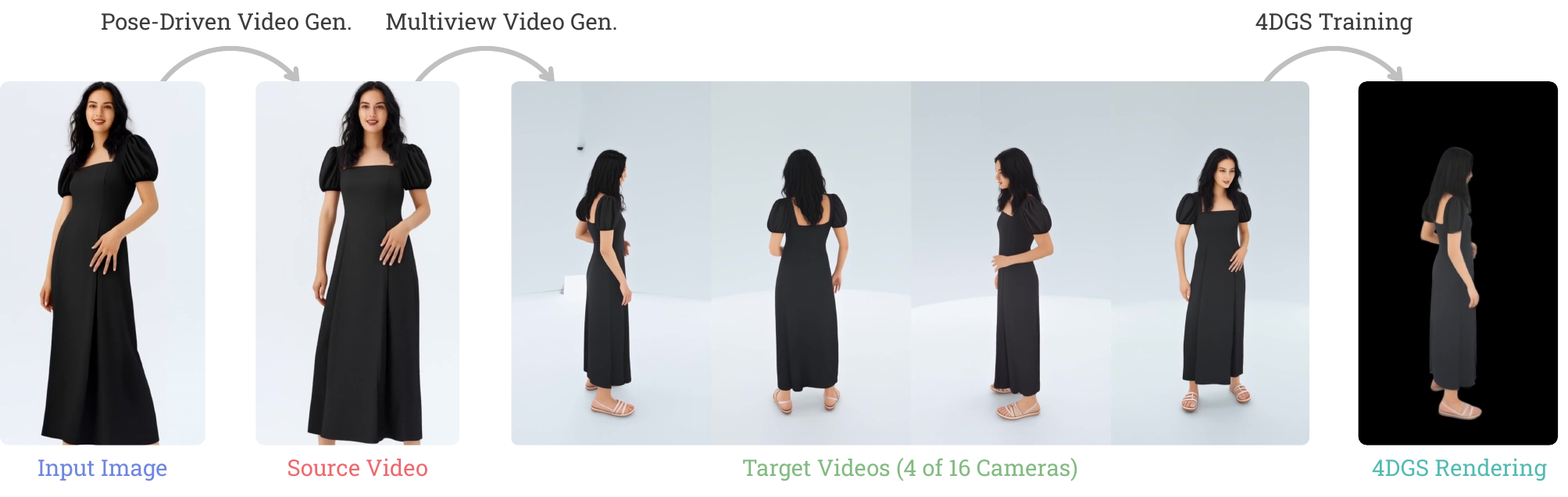}
\caption{\textbf{Single image to 4D avatar.} Given a single input image, we first generate a source video via pose-driven video generation (Wan-Animate), then produce multi-view target videos with \sysname{}, and finally reconstruct a 4DGS avatar via FreeTimeGS.}
\Description{A left-to-right pipeline turns one portrait into a pose-driven source video, generates synchronized target-view videos with 4DAnyone, and reconstructs an animatable 4D Gaussian avatar.}
\label{fig:supp_image2avatar}
\end{figure*}

\PAR{Single image to 4D avatar.}
We can achieve the full single-image-to-4D pipeline by chaining an off-the-shelf animation model (Wan-Animate~\cite{cheng2025wananimate}) with \sysname{}.
Given a single input image and a driving motion video, Wan-Animate first generates a monocular video of the subject performing the target motion.
\sysname{} then takes this synthesized video as input and produces multi-view consistent videos, from which a 4DGS avatar is reconstructed via FreeTimeGS~\cite{wang2025freetimegs}, as illustrated in Fig.~\ref{fig:supp_image2avatar}.

\section{Limitations and Failure Cases}
\label{sec:supp_limitations}

\begin{figure*}[t]
\centering
\includegraphics[width=\textwidth]{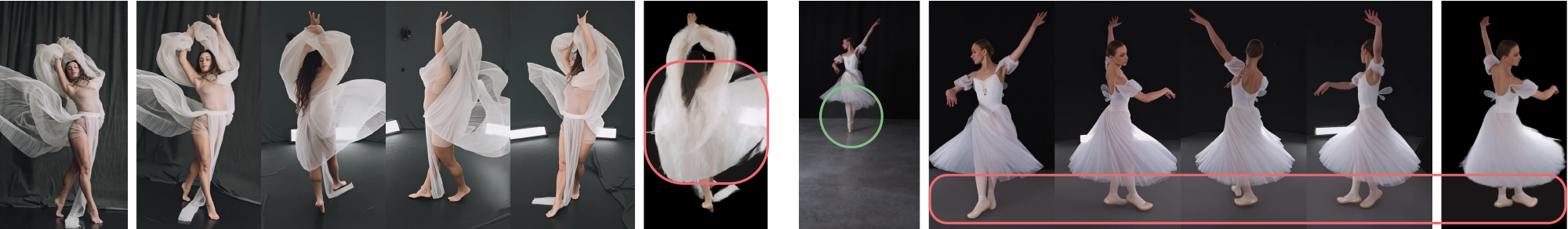}
\caption{\textbf{Failure cases.}
Left: skeleton guidance is uninformative for the large flowing fabric, which is generated inconsistently across views and yields a degraded reconstruction (red box).
Right: HMR mis-estimates the en-pointe pose (green circle in the source view) as flat feet, and all generated views inherit the pose error (red box).}
\Description{Two failure examples. Red boxes highlight inconsistent large flowing fabric in one reconstruction and incorrect feet in generated views caused by a source-pose estimation error marked with a green circle.}
\label{fig:failure_cases}
\end{figure*}

Our skeleton-conditioned pipeline is robust to many challenging inputs: even under occlusion or motion blur, HMR typically still predicts a complete and plausible skeleton, so pose errors yield a slightly shifted yet coherent human rather than a broken result.
Fig.~\ref{fig:failure_cases} shows two representative failure modes.

\PAR{Loose garments.}
Skeleton guidance is not informative for garments that move far from the body.
In Fig.~\ref{fig:failure_cases} (left), the large flowing fabric is generated inconsistently across views, leading to a degraded reconstruction.

\PAR{Inaccurate pose estimation.}
When HMR mis-estimates an unusual pose, the generation faithfully follows the wrong skeleton.
In Fig.~\ref{fig:failure_cases} (right), the dancer stands en pointe in the source video, but HMR predicts flat feet, and all generated views inherit this pose error.

\end{document}